\documentclass[nonacm, sigconf, screen]{acmart}

\usepackage{microtype}
\usepackage{graphicx}
\usepackage{subfigure}
\usepackage{booktabs} % for professional tables

\usepackage{xcolor} % Changing colors
\definecolor{blue_color}{HTML}{038DF5} 
\definecolor{red_color}{HTML}{FF0000}
\definecolor{purple_color}{HTML}{9933FF}
\usepackage{enumitem} % Edit the items in the text
\usepackage{multirow}

\definecolor{darkergreen}{RGB}{74, 160, 44} % This is just an example; adjust the values to get the shade you want

\AtBeginDocument{%
  \providecommand\BibTeX{{%
    \normalfont B\kern-0.5em{\scshape i\kern-0.25em b}\kern-0.8em\TeX}}}

\acmISBN{978-1-4503-XXXX-X/18/06}

\begin{document}

\title{PATE: Proximity-Aware Time series anomaly Evaluation}

\author{Ramin Ghorbani}
\authornote{Corresponding Author: r.ghorbani@tudelft.nl \\ Accepted by ACM KDD 2024, Research Track (Preprint version)}

\affiliation{%
  \institution{Delft University of Technology}
  \city{Delft}
  \country{Netherlands}
}
%\email{r.ghorbani@tudelft.nl}

\author{Marcel J.T. Reinders}
\affiliation{%
  \institution{Delft University of Technology}
  \city{Delft}
  \country{Netherlands}
}

\author{David M.J. Tax}
\affiliation{%
  \institution{Delft University of Technology}
  \city{Delft}
  \country{Netherlands}
}


%% article

\begin{abstract}
Evaluating anomaly detection algorithms in time series data is critical as inaccuracies can lead to flawed decision-making in various domains where real-time analytics and data-driven strategies are essential. Traditional performance metrics assume iid data and fail to capture the complex temporal dynamics and specific characteristics of time series anomalies, such as early and delayed detections. We introduce Proximity-Aware Time series anomaly Evaluation (PATE), a novel evaluation metric that incorporates the temporal relationship between prediction and anomaly intervals. PATE uses proximity-based weighting considering buffer zones around anomaly intervals, enabling a more detailed and informed assessment of a detection. Using these weights, PATE computes a weighted version of the area under the Precision and Recall curve. Our experiments with synthetic and real-world datasets show the superiority of PATE in providing more sensible and accurate evaluations than other evaluation metrics. We also tested several state-of-the-art anomaly detectors across various benchmark datasets using the PATE evaluation scheme. The results show that a common metric like Point-Adjusted F1 Score fails to characterize the detection performances well, and that PATE is able to provide a more fair model comparison. By introducing PATE, we redefine the understanding of model efficacy that steers future studies toward developing more effective and accurate detection models.\\
Public source code: \url{https://github.com/Raminghorbanii/PATE}
\end{abstract}


\begin{CCSXML}
<ccs2012>
   <concept>
       <concept_id>10002944.10011123.10011130</concept_id>
       <concept_desc>General and reference~Evaluation</concept_desc>
       <concept_significance>500</concept_significance>
       </concept>
   <concept>
       <concept_id>10002944.10011123.10011124</concept_id>
       <concept_desc>General and reference~Metrics</concept_desc>
       <concept_significance>500</concept_significance>
       </concept>
   <concept>
       <concept_id>10002944.10011123.10010916</concept_id>
       <concept_desc>General and reference~Measurement</concept_desc>
       <concept_significance>500</concept_significance>
       </concept>
   <concept>
       <concept_id>10002950.10003714</concept_id>
       <concept_desc>Mathematics of computing~Mathematical analysis</concept_desc>
       <concept_significance>500</concept_significance>
       </concept>
 </ccs2012>
\end{CCSXML}

%\ccsdesc[500]{General and reference~Evaluation}
%\ccsdesc[500]{General and reference~Metrics}
%\ccsdesc[500]{General and reference~Measurement}
%\ccsdesc[500]{Mathematics of computing~Mathematical analysis}

\keywords{Time Series, Anomaly Detection, Evaluation Metrics, Precision, Recall}


\maketitle

\vspace{2mm}
\section{Introduction}
\label{se: Introduction}

Anomaly detection in time series (TS) data, the process of identifying unusual patterns that deviate from the expected norm, has become increasingly important across various domains \cite{dif_Anomaly, PPG_Anomaly}. The rapid advancement of data-driven decision-making and real-time analytics has opened opportunities for developing more accurate anomaly detection methods. Such developments often lead to models competing to claim the status of 'State-of-the-Art' (SOTA). Achieving this status is not just a matter of academic prestige; it often directs the focus of future research, influences industry adoption, and guides the development of practical applications. However, choosing an appropriate evaluation metric is critical to avoid incorrect conclusions about a model's performance. Relying on evaluation metrics that do not accurately reflect the true effectiveness of the models can lead to flawed decisions in real-world applications. This is particularly consequential in critical domains, such as medical diagnostics or financial fraud detection, where relying on a poorly evaluated model can have serious repercussions.

\begin{figure}[t]
\begin{center}
\centerline{\includegraphics[width=\columnwidth]{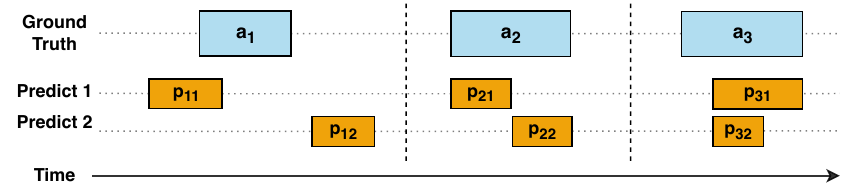}}
\vspace{-1mm}
\caption{\textit{Illustration of anomaly detection in time series data.} \( a_{1-3} \) represent the actual anomalies as ground truth. Predictions are denoted by \( p\). The durations of both events are indicated by the length of the boxes. Overlapping areas between \( p\) and \( a\) demonstrate where the model has correctly identified anomalies.}
\label{Figure_Intro}
\end{center}
\vspace{-5mm}
\end{figure}

Standard evaluation metrics such as Precision and Recall \cite{Precision_Recall_paper} are effective for point-based anomaly detection as they assess the accuracy of detecting isolated iid events. In this context, each data point is evaluated independently, allowing for straightforward calculation of these metrics. However, in TS data, events and anomalies typically occur in time \emph{intervals}. This complexity causes several situations: 1) \textit{Early Detection}, when potential anomalies are identified before they fully manifest, based on subtle changes in the data pattern over time. Figure \ref{Figure_Intro} shows an example of early detection where prediction \( p_{11} \) detects the anomaly event \( a_{1} \) earlier than its actual occurrence. Although \( p_{11} \) does not align exactly with \( a_{1} \), such early detection is valuable for early response actions and should be appropriately appreciated in evaluation metrics. 2) \textit{Delayed Detection}, occurs when an anomaly event is not detected immediately but is identified at a later time, even after its actual occurrence. In Figure \ref{Figure_Intro}, the anomaly event \( a_{1} \) is detected with a delay by prediction event \( p_{12} \). Although \( p_{12} \) does not align precisely with \( a_{1} \), this type of delayed detection should be accounted for in the evaluation process, as it reflects the model's ability to eventually identify anomalies, even after some delay.

Another situation, 3) \textit{Onset Response Time}, refers to how close the detection of an anomaly is to the start of the event. Timely detection is valuable, especially in scenarios where immediate action is required. In Figure \ref{Figure_Intro}, anomaly event \( a_{2} \) is detected by \( p_{21} \) and \( p_{22} \). However, \( p_{21} \)  aligns more closely with the beginning of the anomaly event \( a_{2} \), indicating a faster response than \( p_{22} \). Evaluation metrics should reward those that occur promptly after the onset of an anomaly. Finally 4) \textit{Coverage level of Predictions}, refers to the range that a prediction covers an actual anomaly. The effectiveness of a prediction can be measured by how much of the anomaly it successfully captures. In Figure \ref{Figure_Intro}, predictions \( p_{31} \) and \( p_{32} \) both detect anomaly event \( a_{3} \), but \( p_{31} \) covers \( a_{3} \) more than \( p_{32} \). This more extensive coverage by \( p_{31} \) makes it a more effective prediction for \( a_{3} \). Accordingly, evaluation metrics need to consider the coverage range of the predictions over the duration of the anomalies.

\noindent Various metrics have been developed that are specifically tailored to the sequential nature of time series data (referred to as \textit{Sequential Adaptability}). For instance, Range-based Precision and Recall metrics, hereafter denoted as \textit{R-based}~\cite{range_based_fscore}, expand upon traditional metrics by incorporating factors such as existence (detecting the anomaly range with at least one point), size and position (reflecting the number and relative position of correctly detected anomaly ranges), and cardinality (penalizing fragmented predictions for a single anomaly). The Time Series Aware Precision and Recall, hereafter denoted as \textit{TS-Aware}~\cite{TS_Aware_fscore}, follows a similar approach but omits cardinality and position considerations. This metric requires a prediction to cover a minimum percentage \( \theta \) of an anomaly for it to be considered a true detection. They also add a buffer zone \(\delta\) to give some credit for delayed detection in a decreasing manner. An enhanced version, denoted as \textit{ETS-Aware}~\cite{E_TS_Aware_fscore}, further refines the evaluation by combining detection and overlap scores for improved accuracy in scoring overlapped detections. Further, the \textit{Affiliation} metric~\cite{Affiliation_fscore}, introduces a different perspective by focusing on the distance between prediction and actual anomaly ranges. It assesses the proximity of predicted anomalies to actual ones by measuring the duration between their respective ranges. 

Another widely used method is the Point Adjusted F1 Score metric, which we will denote as \textit{PA-F1}~\cite{PA}. This approach assumes that detecting a single point in an anomaly range is sufficient for human experts to identify the entire range. Thus, it considers all observations within the corresponding anomaly range as correctly detected anomalies. However, it has been criticized for potentially generating optimistic scores. For example, \cite{PA_K} revealed that random anomaly scores from a uniform distribution outperform state-of-the-art methods when evaluated using this metric. To address this, \cite{PA_K} proposed a modified version that requires a portion of \( K\% \) of the anomaly range to be detected before making any adjustments. 

While all these metrics represent advancements in time series anomaly detection evaluation, they do not fully consider all the critical factors of early and delayed detections, or onset response timing. In addition to these limitations, the aforementioned metrics also require the setting of a threshold, a value where data points with anomaly scores exceeding this value are classified as anomalies. Selecting this threshold adds additional complexity and leads to subjectivity and inconsistency in evaluations. Metrics such as the Area Under the Receiver Operating Characteristic curve (AUC-ROC) and the Area Under the Precision-Recall curve (AUC-PR) eliminate the need for thresholding by evaluating the performance of the model across a range of thresholds. However, they fall short in time series contexts due to not considering the order of the data points and the temporal correlation between them. In response to this issue, Volume Under the Surface (VUS) metrics, \textit{VUS-ROC} and \textit{VUS-PR}, are proposed \cite{VUS}. These metrics acknowledge the need to accommodate close predictions to the true anomaly ranges by adjusting the labels to be between 0 and 1 on a range over both sides of the actual anomaly range. Although the method is threshold-free, it does not pay attention to early and delayed detection, and onset response time. Furthermore, by changing the original labels, the metric gives unrealistic scores, as reaching the maximum detection score of 1 is not possible.

\begin{table}[t]
\caption{\textit{Comparison of Anomaly Detection Evaluation Metrics.} Key features: Sequential Adaptability (SA); Early Detection (ED);  Delayed Detection (DD); Onset Response Time (ORT); Coverage Level (CL) and Threshold-Free (TF)}
\vspace{-2mm}
\label{Table_Intro}
\begin{center}
%\small
\fontsize{8.3pt}{10pt}\selectfont % Custom font size
\setlength{\tabcolsep}{6pt} % Adjust the space between columns
\begin{tabular}{lccccccc}
\toprule
Metric & SA & ED & DD & ORT & CL & TF  \\
\midrule
Precision/Recall (F1 Score) & - & - & - & - & - & -  \\
R-based  & \checkmark & - & - & - &  - &  - \\
TS-Aware/ETS-Aware  & \checkmark & - & \checkmark & - &  - & -  \\
Affiliation  & \checkmark & - & - & - &  \checkmark & - \\
PA-F1  & \checkmark & - & - & - &  - & -  \\
AUC-ROC/PR & - & - & - & - &  - & \checkmark \\
VUS-ROC/PR & \checkmark & - & \checkmark & - &  - & \checkmark \\
\textbf{PATE} & \checkmark & \checkmark & \checkmark & \checkmark &  \checkmark & \checkmark  \\
\bottomrule
\end{tabular}
\end{center}
\vspace{-2mm}
\end{table}

This paper introduces a novel evaluation metric named the Proximity-Aware Time series anomaly Evaluation (PATE) method. Our approach integrates buffer zones around the anomaly events and utilizes a special proximity-based weighting mechanism, enabling a detailed assessment of both early/delayed detections and addressing the onset response time challenge. PATE avoids the subjectivity of threshold-dependent metrics by integrating over the range of thresholds, offering a fair and unbiased evaluation, especially in research settings where expert knowledge might
not be available for setting the exact desirable parameters based on the application. Table \ref{Table_Intro} illustrates a comparison between existing metrics and PATE, highlighting the comprehensive adaptability reconsideration of PATE in evaluating the TS anomaly detection.

\begin{figure*}[t]
%\vskip 0.2in
\centering % Center the figure
\includegraphics[width=\textwidth]{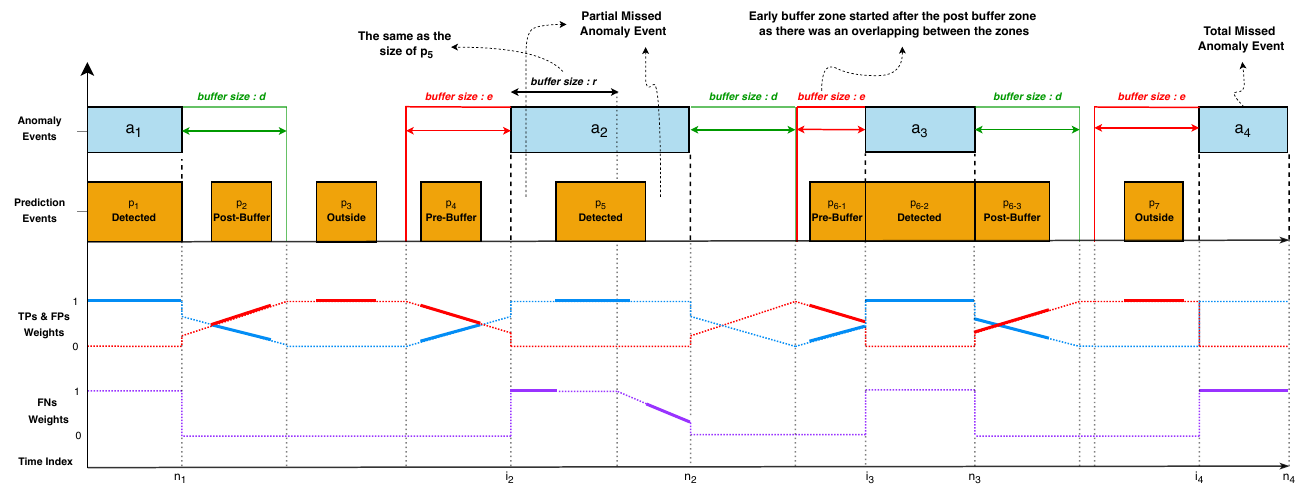} 
\vspace{-7mm}
\caption{\textit{Illustration of the Categorization and Weighting Mechanism in the PATE Method.} Prediction events (\( p_{1}-p_{7} \)) are represented by orange boxes, while anomaly events (\( a_{1}-a_{4} \)) are depicted by blue boxes. TP weights are illustrated with a blue line \textcolor{blue_color}{\rule{0.5cm}{0.1cm}}, FP weights with a red line \textcolor{red_color}{\rule{0.5cm}{0.1cm}}, and FN weights with a purple line \textcolor{purple_color}{\rule{0.5cm}{0.1cm}}. Note that the solid segments of the lines, in contrast to the dotted segments, indicate the activated weights for the example scenario depicted in the figure.}

\label{fig:categorization}

\end{figure*}

\section{Proposed Evaluation Metric - PATE}

A time series is denoted as a sequence of observations \( \mathcal{\boldsymbol{X}} = \left\{ {x}_{t} \right\}_{t=1}^{T} \), where \( T \) represents the length of the time series, and each \( {x}_t \) is the observed data point at time \( t \).

An actual anomaly event (labeled as positive in the ground truth labels) is a subsegment within the time series, denoted as \( \boldsymbol{a}_k =  ({i_k}, {n_k}) \) for points \( i_k \) and \( n_k \) with \( 1 \leq i_k \leq n_k \leq T \). The set of all anomaly events in the time series is represented as \( \mathcal{\boldsymbol{A}} = \left\{ \boldsymbol{a}_k \right\}_{k=1}^{N} \), where \( N \) is the number of anomaly events present in the time series. 

In practice, the detection models output continuous anomaly scores, denoted as \( \mathcal{\boldsymbol{S}} = \left\{ {s}_{t} \right\}_{t=1}^{T} \), representing the likelihood of each observation \( {x}_t \) to be anomalous. These scores are then converted into binary predictions by applying a threshold \(\theta\), where scores equal to or exceeding the threshold are classified as anomalies. We define a prediction event as a subsegment identified by these binary predictions to be anomalous, denoted as \( \boldsymbol{p}_l(\theta) =  ({m_l}, {j_l}) \) for points \( m_l \) and \( j_l \) with \( 1 \leq m_l \leq j_l \leq T \). The set of all prediction events is represented as \( \mathcal{\boldsymbol{P}} = \left\{ \boldsymbol{p}_l(\theta) \right\}_{l=1}^{M} \), where \( M \) is the number of prediction events identified by the model.

The effectiveness of the anomaly detection model is determined by how well these \( \boldsymbol{p}_l(\theta) \) events align with the \( \boldsymbol{a}_k \) events. PATE distinguishes several categories of matches between ground truth and predictions based on their temporal relationships and assigns proximity-specific weights to each point in each category. These weights are then used to compute a weighted version of Precision and Recall scores. The final measure of PATE is a weighted AUC-PR, which is derived from these weighted Precision and Recall scores. Further details on these computations are provided in the following sections.

\subsection{Categorizing the Events}
\label{se: Categories}
Figure \ref{fig:categorization} illustrates the different categories of anomaly and prediction events in relation to each other. In assessing each \( \boldsymbol{p}_l(\theta) \), we consider its overlap, proximity, or distance (temporal relation) from each \( \boldsymbol{a}_k \). This approach allows for the clear differentiation of the diverse scenarios: complete and partial detection of anomalies, early or delayed detection, and instances where anomalies are either partially or entirely missed. Specifically, we categorize the anomaly and prediction events as follows:

%\begin{itemize}[leftmargin=*] 

\subsubsection{Prediction events categories:}

\

\vspace{2.2mm}
\textbullet\, \textbf{\textit{True-Detection}}: Sub-segments of the prediction event \( \boldsymbol{p}_l(\theta) \) that overlap with an anomaly event \( \boldsymbol{a}_k \), indicating anomalies that are accurately identified and not missed. Examples are segments \( p_{1} \), \( p_{5} \), and \( p_{6-2} \) in Figure \ref{fig:categorization}. % Detected is Actual in the code. 

\

\textbullet\, \textbf{\textit{Post-Buffer Detection}}: Sub-segments of the prediction event \( \boldsymbol{p}_l(\theta) \) that fall into a buffer zone immediately following an anomaly event \( \boldsymbol{a}_k \) (See segments \( p_{2} \) and \( p_{6-3}\) in Figure \ref{fig:categorization}). This category highlights the capacity of the model for delayed detection. The post-buffer zone size, denoted by \( d \), can be adjusted by experts based on specific application needs. When \( d \) is unknown for a specific application, we can consider a range of values for \( d \) rather than a fixed one \( D = \{0, 1, \ldots, d_{\max}\} \). This approach allows for a comprehensive assessment of the model's performance across different scenarios, as each buffer size can provide a different perspective on the performance of the model. Details on how these buffer sizes contribute to the overall PATE score will be discussed in the following sections. 

\

\textbullet\, \textbf{\textit{Pre-Buffer Detection}}: Sub-segments of the prediction event \( \boldsymbol{p}_l(\theta) \) that fall into a zone that precedes the start of an anomaly event \( \boldsymbol{a}_k \). This category highlights the capacity of the model for early detection, signaling potential anomalies ahead of time.  Similar to the post-buffer zone, the size of the pre-buffer zone, denoted by \( e \), varies within the set \( E = \{0, 1, \ldots, e_{\max}\} \) with the same approach for the assessment. The assignment of points to this category is conditional on not overlapping with the Post-Buffer zone of a preceding anomaly \( \boldsymbol{a}_{k-1} \), ensuring that the model early warning is distinct from a delayed detection of the previous event. In other words, the Post-Buffer category has priority, and therefore, if \( i_k - e < n_{k-1} + d \) then the Pre-Buffer zone starts at \(n_{k-1} + d + 1\) instead of \( i_k - e\). Furthermore, Pre-Buffer detection is dependent on the successful detection of the subsequent anomaly event \( \boldsymbol{a}_k \). In situations where no part of the subsequent event \( \boldsymbol{a}_k \) is detected by a True-Detection, this Pre-Buffer detection is considered a false alarm rather than a meaningful early detection.  Consequently, this early prediction \( \boldsymbol{p}_l(\theta) \) is reclassified as False Positive (the Outside category, which is discussed below). Further details are given in Appendix~\ref{Clarification_DD_ED}. In Figure \ref{fig:categorization}, \( p_{4} \) and \( p_{6-1}\) are the examples of pre-buffer detection category, whereas \( p_{7} \) is not considered in this category.

\

\textbullet\, \textbf{\textit{Outside}}: Sub-segments of the prediction event \( \boldsymbol{p}_l(\theta) \) located outside the ranges of anomaly event \( \boldsymbol{a}_k \) and its buffer zones. These are instances where the model incorrectly flags normal behavior as anomalous (False Positive), like segments \( p_{3} \) and \( p_{7}\) in Figure \ref{fig:categorization}.

\subsubsection{Anomaly events categories:}

\

\vspace{2.2mm}

\textbullet\, \textbf{\textit{Total Missed Anomalies}}: When an entire anomaly event \( \boldsymbol{a}_k \) is not detected by any segments of the prediction event \( \boldsymbol{p}_l(\theta) \), that is, all detections are before \( i_k - e\) or after \( n_k + d\). This category indicates a complete failure (False Negative) of the model to identify the anomaly. See segment \( a_{4} \) in Figure \ref{fig:categorization}. 

\

\textbullet\, \textbf{\textit{Partial Missed Anomalies}}: This category is assigned when only a part of anomaly event \( \boldsymbol{a}_k \) is detected by the prediction events \( \boldsymbol{p}_l(\theta) \)'s, but there are segments within the anomaly range of \( \boldsymbol{a}_k \) that remain undetected. This category not only highlights the model's capability to detect parts of an anomaly but also its inability to identify the anomaly event in its entirety. For instance, segment \( a_{2} \) in Figure \ref{fig:categorization}, where a part of it is detected by \( p_{5} \) but before and after \( p_{5} \) we have partially missed segments.

%\end{itemize}

\vspace{2mm}
\subsection{Weighting Process}

After each individual time point is assigned to its category, we define weights for each of these points to determine their contribution to the True Positive (TP), False Positive (FP), and False Negative (FN) metrics of the detector. It is important to note that time points at which no anomaly is present and no prediction is made, True Negatives (TN), do not actively contribute to the performance metrics and are, therefore, implicitly assigned a weight of zero, reflecting their non-contribution. The bottom half of Figure \ref{fig:categorization} visually represents the variations in weights across all different categories.

%\begin{itemize} [leftmargin=*] 

\

\textbullet\, \textbf{\textit{True-Detection Weights:}}
Each point \( {t} \) from the True-Detection category, lying within the range of an anomaly event \([{i_k}, {n_k}]\), is considered correctly identified. Thus, such points are assigned the maximum weight of 1 as True Positives:

\begin{equation}
\small
w^{\text{TP}}(t) = 1 \quad \text{for } t \in TrueDetection ~{\boldsymbol{p}_l(\theta)}
\label{eq:True_Detection_weight}
\end{equation}

\

\textbullet\, \textbf{\textit{Post-Buffer Detection Weights:}}
Each point \( {t} \) from the post-buffer category, in the range of \(({n_k}, {n_k}+d]\), is evaluated in relation to the anomaly event \( \boldsymbol{a}_k \). These points, while not being true positives in the traditional sense, receive a weight based on their proximity to the \( \boldsymbol{a}_k \), which captures the diminishing influence of an anomaly over time as the distance from the anomaly event increases.

\begin{equation}
\small
w^{\text{TP}}(t) = 1 - \frac{\sum_{y={i_k}}^{n_k} |{t} - y|}{\sum_{y=i_k}^{n_k} |(n_k + d) - y|} \quad \text{for } t \in \text{Post-Buffer}~{\boldsymbol{p}_l(\theta)}
\label{eq:post_buffer_TP_weight}
\end{equation}

\

Here, the numerator calculates the distance of \( {t} \) from each point within the anomaly event, and the denominator normalizes this against the total potential spread within the buffer zone. With this method, we account for the proximity to the entire anomaly, not just its endpoint. Thus, we address the delayed detection by recognizing that any point within the actual anomaly range might influence predictions in the buffer zone, not just the most immediate or final points of the anomaly. This also implies that the lengths of the anomalies influence the weights. For smaller anomalies, points in the Post-Buffer zone are closer to the anomaly onset, and will therefore be assigned with higher true positive weights. Further details, regarding the impact of anomaly length on the weights, are given in Appendix~\ref{Effect_lenght_Ws}.

In the Post-Buffer zone, as the distance from \( \boldsymbol{a}_k \) increases, the likelihood of a detection being a False Positive rises. Thus, the weights assigned to false positives in this zone are calculated as the complement of the TPs weights, acknowledging the reduced significance of detections further from the actual anomaly. Figure \ref{fig:categorization} visually shows the variations in TP and FP weights across the Post-Buffer categories (\( p_{2} \) and \( p_{6(3)} \)). 

\begin{equation}
\small
w^{\text{FP}}(t) = 1 - w^{\text{TP}}(t) \quad \text{for } t \in \text{Post-Buffer}~{\boldsymbol{p}_l(\theta)}
\label{eq:post_buffer_FP_weight}
\end{equation}

\

\textbullet\, \textbf{\textit{Outside Weights:}}
Each point \( {t} \) from the Outside category indicates a situation where the model incorrectly identifies normal behavior as anomalous. Given the lack of proximity to any real anomaly, these points are considered FPs with a maximum weight of 1, reflecting a significant deviation from accurate detection.

\begin{equation}
\small
w^{\text{FP}}(t) = 1 \quad \text{for } t \in \text{Outside}~{\boldsymbol{p}_l(\theta)}
\label{eq:Outsie_weight}
\end{equation}

\

\textbullet\, \textbf{\textit{Pre-Buffer Detection Weights:}}
Each point \( {t} \) in the pre-buffer category, in the range of \([i_k - e, i_k)\), is assessed for potential early detection in relation to the preceding \( \boldsymbol{a}_k \). These points, while not being true positives in the conventional sense, are evaluated for their proximity to the upcoming anomaly:

\begin{equation}
\small
w^{\text{TP}}(t) = 1 - \frac{\sum_{y=i_k}^{n_k} |y - {t}|}{\sum_{y=i_k}^{n_k} |(i_k - e) - y|} \quad \text{for } t \in \text{Pre-Buffer}~{\boldsymbol{p}_l(\theta)}
\label{eq:Pre-Buffer_TP_weight}
\end{equation}

\

Here, the numerator represents the distance of \( {t} \) from every point in \( \boldsymbol{a}_k \), capturing how early \( {t} \) occurs relative to the anomaly. The denominator provides normalization against the total potential spread within the pre-buffer zone. This mechanism recognizes that any point within the anomaly event might have an influence on the zone. 

Similar to the Post-Buffer zone, the likelihood of a point being a False Positive increases as the distance from the \(i_k\) increases. Thus, the weights assigned to FPs are calculated as the complement of the TPs weights, reflecting the reduced relevance of premature detections. Figure \ref{fig:categorization} shows the variations in weights of the Pre-Buffer categories (\( p_{4} \) and \( p_{6(1)} \)). 

\begin{equation}
w^{\text{FP}}(t) = 1 - w^{\text{TP}}(t) \quad \text{for } t \in \text{Pre-Buffer}~{\boldsymbol{p}_l(\theta)}
\label{eq:Buffer_FP_weight}
\end{equation}

\

\textbullet\, \textbf{\textit{Total Missed Anomalies Weights:}}
When the entire range of \( \boldsymbol{a}_k \) is undetected, each  \( {t} \) within its interval receives a maximum False Negative weight of 1. This assignment underscores the complete failure of the model in detecting the anomaly event. The variations in FN weight across \( a_{4} \) as a total missed event are shown in Figure \ref{fig:categorization}.

\begin{equation}
w^{\text{FN}}(t) = 1 \quad \text{for } t \in \text{Total-Missed}~{\boldsymbol{a}_k}
\label{eq:Total_Missed_weight}
\end{equation}

\

\textbullet\, \textbf{\textit{Partial Missed Anomaly Weights:}}
When \( \boldsymbol{a}_k \) is only partially detected, the undetected points \( {t} \) within \( \boldsymbol{a}_k \), are evaluated based on their proximity to the start of the anomaly event. The closer the points are to the anomaly onset the higher the FN weight, emphasizing the onset response time in detection. Here $\text{for } t \in \text{\textit{Partial Missed}}~{\boldsymbol{a}_k}$, we have:

\begin{equation}
\small
w^{\text{FN}}(t) = \begin{cases} 
    1 & \text{if } {t} \leq i_k + r \\
    1 - \frac{\sum_{y=i_k}^{i_k + r} |{t} - y|}{\sum_{y=i_k}^{n_k} |n_k - y|} & \text{otherwise} 
    \end{cases} 
\label{eq:PartialMissed_weight}
\end{equation}

\

Here, \( r \) is the size of the buffer that starts from the onset of the anomaly event. Undetected points in this buffer are penalized with a maximum FN weight of 1. Undetected points outside the buffer received a reduced
FN weight, weighted by the distance to the buffer. The rationale behind this design is that more comprehensive coverage of an anomaly by a prediction justifies a more lenient assessment of its exact timing accuracy. In other words, when a prediction successfully captures a larger portion of \( \boldsymbol{a}_k \), the precision of its onset timing becomes less critical. Therefore, \( r \) is defined as the fraction of coverage of \( \boldsymbol{a}_k \) by its corresponding \( \boldsymbol{p}_l (\theta) \). Figure \ref{fig:categorization} shows the variations in FN weight across the Partial Missed category where some segments of \( a_{2} \) are missed.

%\end{itemize}

\subsection{PATE Final Score}
\label{se:PATE}

The PATE final metric is designed to comprehensively evaluate anomaly detection by considering a full range of combinations of pre-buffer (\(e\)) and post-buffer (\(d\)) sizes. For each combination of \(e\) and \(d\), we apply a range of thresholds (\(\theta\)) to convert the continuous anomaly scores (\(\mathcal{\boldsymbol{S}}\)) into binary predictions, capturing the model's performance across different sensitivity levels. Based on these binary predictions, we identify the prediction events \( \mathcal{\boldsymbol{P}}\) and then categorize all prediction and anomaly events. Based on this categorization, we assign appropriate weights to each observation.

We calculate weighted Precision and Recall across all thresholds in the considered range for each specific combination of \(e\) and \(d\). Using these calculations, we construct the Precision-Recall curve for each combination and compute the area under the curve (AUC-PR). Note that the weights $w^{\text{TP}}(t)$, $w^{\text{FP}}(t)$, and $w^{\text{FN}}(t)$ are assigned based on the categorization of each time point $t$. For time points that do not fall into any specific category, the weights are considered to be 0. Thus, the summation in the formulas for Precision and Recall effectively includes only those time points that have been categorized.

\begin{equation}
\text{Precision}_{e,d}(\theta) = \frac{\sum_{t=1}^{T} w^{\text{TP}}({t})}{\sum_{t=1}^{T} w^{\text{TP}}({t}) + \sum_{t=1}^{T} w^{\text{FP}}({t})}
\label{eq:precision}
\end{equation}

\begin{equation}
\text{Recall}_{e,d}(\theta) = \frac{\sum_{t=1}^{T} w^{\text{TP}}({t})}{\sum_{t=1}^{T} w^{\text{TP}}({t}) + \sum_{t=1}^{T} w^{\text{FN}}({t})}
\label{eq:recall}
\end{equation}

\

Finally, the overall PATE score is determined by averaging the computed AUC-PRs across all combinations of \(e\) and \(d\):

\begin{equation}
\text{PATE} = \frac{1}{|E| \times |D|}{\sum_{e \in E, d \in D} \text{AUC-PR}_{e,d}}
\label{eq:PATE}
\end{equation}

\

Here, \( |D| \) and \( |E| \) represent the number of distinct values for \( d \) and \( e \) within their respective sets.

\section{Experiments and Results}

\subsection{Synthetic Data Experiments}
To highlight the merits of PATE, we first compare PATE with alternative evaluation metrics on a synthetic time series with a binary anomaly detector. The alternative measures can be threshold-dependent or independent. Threshold-independent metrics are inherently evaluated across a range of possible thresholds. For this example, we consider thresholds \(\theta = \{0, 1\}\) to distinguish between normal and anomalous predictions. For threshold-dependent metrics, we define the optimal threshold as \(\theta = 1\), identifying points predicted as '1' (anomalous) for evaluation.

\begin{figure}[!b]
\vspace{-2mm}
\centering
\includegraphics[width=\columnwidth]{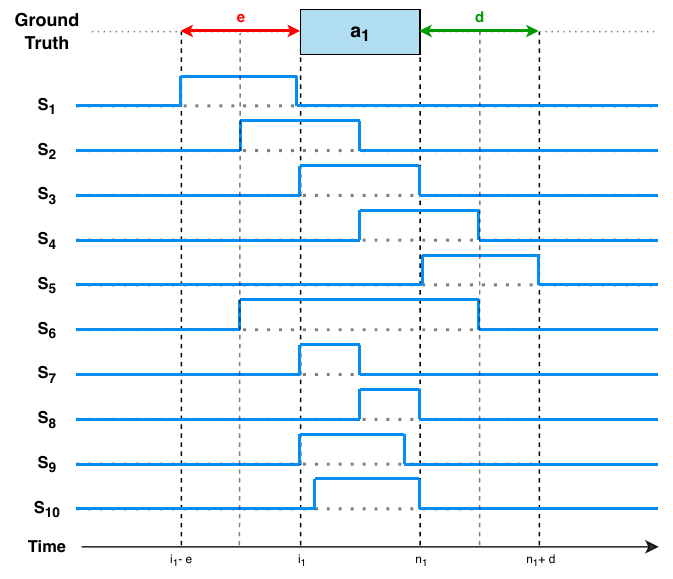}
\vspace{-7mm}
\caption{\textit{Illustration of examples with synthetic data}. The figure shows the placement of different anomaly scores $S$ from a binary anomaly detector.}
\label{fig:scenarios}

\vspace{3mm}

\captionof{table}{\textit{Comparison of evaluation metrics for synthetic data examples depicted in Figure \ref{fig:scenarios}}. 'F1' refers to the F1 Score. 'Standard-F1' specifically denotes the conventional F1 Score calculated from standard Precision and Recall.}

\vspace{-2.5mm}
\label{table:metric-comparison}
\fontsize{8.3pt}{10pt}\selectfont % Custom font size
\setlength{\tabcolsep}{4pt} % Adjust the space between columns
%\small % or \footnotesize or \scriptsize
\begin{tabular}{c|ccccc|ccccc}
\toprule

 &\multicolumn{5}{c|}{\begin{tabular}{@{}c@{}} Threshold-independent \\ Metrics \end{tabular}} & \multicolumn{5}{c}{\begin{tabular}{@{}c@{}} Threshold-dependent \\ Metrics \end{tabular}} \\
\cmidrule(lr){2-6} \cmidrule(lr){7-11}
\rotatebox{90}{Scenarios} & \rotatebox{90}{PATE} & \rotatebox{90}{VUS-ROC} & \rotatebox{90}{VUS-PR} & \rotatebox{90}{AUC-ROC} & \rotatebox{90}{AUC-PR} & \rotatebox{90}{Standard-F1} & \rotatebox{90}{PA-F1} & \rotatebox{90}{R-based-F1} & \rotatebox{90}{ETS-Aware-F1} & \rotatebox{90}{Affiliation-F1} \\
\midrule

$S_{1}$ & $\textbf{0.03}$ & $0.63$ & $0.37$ & $0.48$ & $0.02$ & $0.00$ & $0.00$ & $0.00$ & $0.00$ & $0.94$ \\
$S_{2}$ & $\textbf{0.76}$ & $0.79$ & $0.72$ & $0.74$ & $0.51$ & $0.50$ & $0.80$ & $0.60$ & $0.75$ & $0.98$ \\
$S_{3}$ & $\textbf{1.00}$ & $0.87$ & $0.88$ & $1.00$ & $1.00$ & $1.00$ & $1.00$ & $1.00$ & $1.00$ & $1.00$ \\
$S_{4}$ & $\textbf{0.69}$ & $0.79$ & $0.70$ & $0.74$ & $0.51$ & $0.50$ & $0.80$ & $0.60$ & $0.75$ & $0.98$\\
$S_{5}$ & $\textbf{0.31}$ & $0.63$ & $0.34$ & $0.48$ & $0.02$ & $0.00$ & $0.00$ & $0.00$ & $0.00$ & $0.94$ \\
$S_{6}$ & $\textbf{0.87}$ & $0.99$ & $0.91$ & $0.98$ & $0.75$ & $0.67$ & $0.67$ & $0.75$ & $0.86$ & $0.98$\\
$S_{7}$ & $\textbf{0.85}$ & $0.69$ & $0.71$ & $0.75$ & $0.76$ & $0.67$ & $1.00$ & $0.75$ & $0.86$ & $0.99$\\
$S_{8}$ & $\textbf{0.77}$ & $0.69$ & $0.71$ & $0.75$ & $0.76$ & $0.67$ & $1.00$ & $0.75$ & $0.86$ & $0.99$ \\
$S_{9}$ & $\textbf{0.95}$ & $0.78$ & $0.79$ & $0.88$ & $0.88$ & $0.86$ & $1.00$ & $0.89$ & $0.93$ & $1.00$ \\
$S_{10}$ & $\textbf{0.88}$ & $0.78$ & $0.79$ & $0.88$ & $0.88$ & $0.86$ & $1.00$ & $0.89$ & $0.93$ & $1.00$ \\

\bottomrule
\end{tabular}
\end{figure}

Figure\ref{fig:scenarios} shows anomaly $a_1$ with its pre and post-buffer zones. Below, ten different detection scenarios are shown, $S_{1}, \dots, S_{10}$. Results in Table \ref{table:metric-comparison} demonstrate that PATE effectively distinguishes the scenarios based on temporal proximity, duration, coverage level, and response timing. For instance, although $S_{1}$ is temporally close to the anomaly event, it fails to detect any part of it. In the context of time series, where past data is crucial for prediction, the inability to detect any part of the anomaly after it starts suggests that the prediction might be a true false alarm rather than a meaningful early detection. A low score for $S_{1}$ reflects a metric that appropriately penalizes lucky guesses or irrelevant detections. On the other hand, $S_{2}$ gets a higher score as it captures part of the anomaly itself, and then the non-overlapping part can be recognized as relevant early detection, which should be valued. Note that the PATE score of 0.03 for $S_{1}$ is not exactly zero because it considers a range of thresholds, including zero. At a threshold of 0, every point is labeled as a potential anomaly, thus increasing both true and false positives. This broad consideration prevents the PATE score from being zero for this specific example.

Meanwhile, $S_{1}$ and $S_{2}$ should be evaluated differently from delayed detections $S_{4}$ and $S_{5}$. Although $S_{4}$'s coverage level is the same as that of $S_{2}$, due to response timing, it gets a lower score. Similarly, the evaluation of $S_{5}$ is completely different from $S_{1}$ as it occurs after the anomaly event. This late detection might indicate that the model is responding to the anomaly, albeit with a significant delay. Hence, it is reasonable to evaluate $S_{5}$ higher than $S_{1}$ as it could reflect some response to the actual anomaly, even though it is late and fails to detect any part of the anomaly. Other metrics, while effective in certain scenarios, do not distinguish between the finer details of anomaly detection. For instance, these metrics just mirror the results of $S_{1}$ and $S_{2}$ for $S_{4}$ and $S_{5}$ without considering the early and delayed context. Moreover, $S_{3}$, as an example of accurate detection, is expected to get the maximum score of 1 by all evaluation metrics, and $S_{6}$ is expected to get a lower score than $S_{3}$. However, the VUS-ROC/PR metrics fail to evaluate these scenarios correctly. The scenarios $S_{7}$, $S_{8}$, $S_{9}$, and $S_{10}$ further exemplify the importance of the coverage level and response timing in detection. In each pair, $S_{7}$ and $S_{9}$ detect the anomaly right from the start; thus they should get scored higher than $S_{8}$ and $S_{10}$. While other metrics tend to score these pairs similarly, PATE recognizes the earlier detections in $S_{7}$ and $S_{9}$ and gives them higher scores. Moreover, in scenarios like $S_{9}$ and $S_{10}$, where the anomaly is covered more extensively, PATE assigns less penalties for response timing inaccuracies. This is seen in the smaller score difference between early and late detections in scenarios with greater coverage.

\vspace{-3mm}
\subsection{Real-World Data Experiments}

To validate the practicality and effectiveness of PATE in real-world applications, we extracted some examples from the publicly available and widely used datasets, UCR-KDD21~\cite{KDD_Dataset} and MIT-BIH Arrhythmia (MBA) ECG~\cite {ECG_Data}. The goal is to evaluate how well PATE, alongside other evaluation metrics, distinguishes between various detection models. To ensure a fair comparison, we compare PATE with threshold-independent evaluation metrics, guaranteeing an unbiased comparison of metrics performances.

\begin{figure}[!b]
\vspace{-2mm}
\centering
\includegraphics[width=\columnwidth]{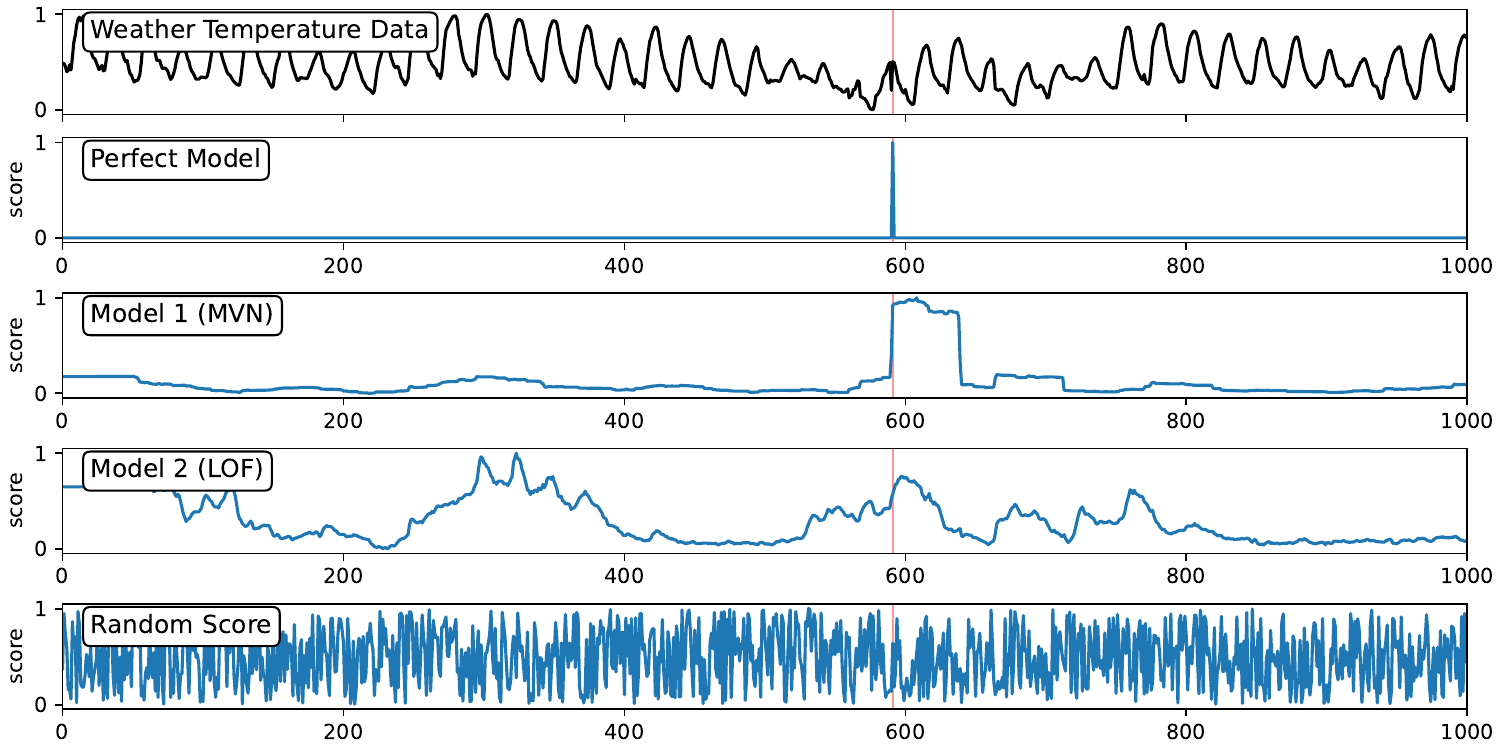} \\
\textit{(a) Weather Temperature data example}
\label{fig:ECG_Example_a}

\vspace{5mm} % Adjust this space as needed

\includegraphics[width=\columnwidth]{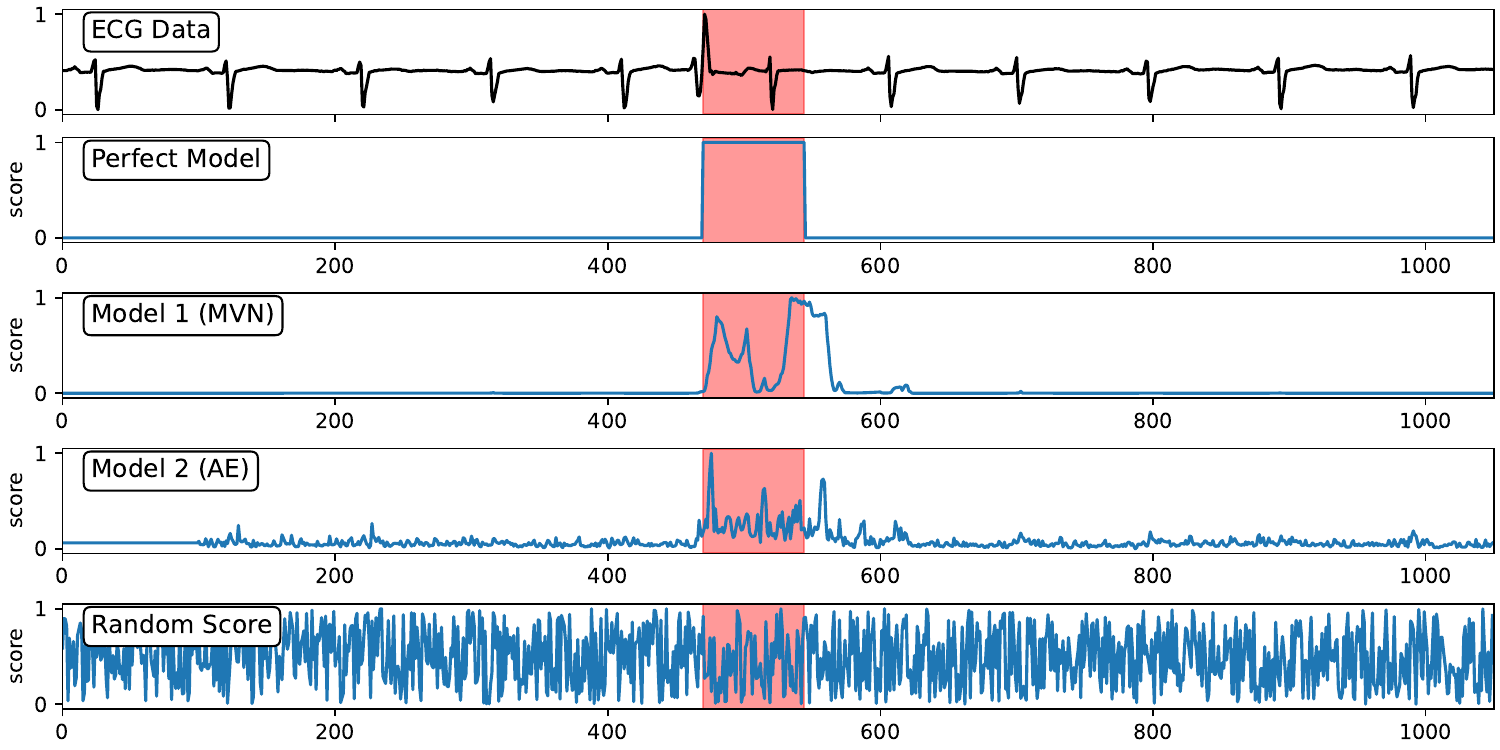}\\
\textit{(b) ECG data example.}
\label{fig:ECG_Example_b}

\caption{\textit{Real-World Datasets and Anomaly Scores of Different Models}. The anomalous segment and its corresponding region (labeled by an expert), against which the models' predictions are compared, is highlighted in red}
\label{fig:Real_Examples}

\vspace{3mm}

\captionof{table}{\textit{Quantitative Evaluation of Anomaly Detection Models}. Evaluation score for different anomaly detection models in detecting the anomalous region in examples of Figure \ref{fig:Real_Examples}.} 

\label{table:real-data-results}
\vspace{-2mm}
\fontsize{8pt}{10pt}\selectfont % Custom font size
\setlength{\tabcolsep}{2.8pt} % Adjust the space between columns
%\small % or \footnotesize or \scriptsize
\begin{tabular}{c|ccccc|ccccc}
\toprule
 Datasets &\multicolumn{5}{c|}{Weather Temperature} & \multicolumn{5}{c}{ECG} \\
\cmidrule(lr){2-11} \cmidrule(lr){7-11}
\rotatebox{0}{Scenarios} & \rotatebox{90}{PATE} & \rotatebox{90}{VUS-ROC} & \rotatebox{90}{VUS-PR} & \rotatebox{90}{AUC-ROC} & \rotatebox{90}{AUC-PR} & \rotatebox{90}{PATE} & \rotatebox{90}{VUS-ROC} & \rotatebox{90}{VUS-PR} & \rotatebox{90}{AUC-ROC} & \rotatebox{90}{AUC-PR} \\
\midrule

{Perfect Model} & $\textbf{1.00}$ & $0.55$ & $0.57$ & $1.00$ & $1.00$             & $\textbf{1.00}$ & $0.90$ & $0.91$ & $1.00$ & $1.00$ \\
{Model 1} & $\textbf{0.88}$ & $0.98$ & $0.71$ & $0.98$ & $0.02$                 & $\textbf{0.83}$ & $0.99$ & $0.89$ & $0.98$ & $0.69$ \\
{Model 2} & $\textbf{0.07}$ & $0.86$ & $0.14$ & $0.83$ & $0.01$                 & $\textbf{0.79}$ & $0.98$ & $0.81$ & $0.97$ & $0.69$ \\
{Random Score} & $\textbf{0.02}$ & $0.67$ & $0.08$ & $0.66$ & $0.01$            & $\textbf{0.07}$ & $0.56$ & $0.11$ & $0.43$ & $0.06$\\

\bottomrule
\end{tabular}

\end{figure}

\begin{figure*}[!ht]
\centering
\captionof{table}{\textit{Comparison of SOTA anomaly detection model using different evaluation metrics across various benchmark datasets.}}
\vspace{-3mm}
\label{table:SOTA_Example}
%\small % or \footnotesize or \scriptsize
\fontsize{8pt}{10pt}\selectfont % Custom font size
\setlength{\tabcolsep}{2.4pt} % Adjust the space between columns

\begin{tabular}{c|cccccc|cccccc|cccccc|cccccc}
\toprule
Datasets & \multicolumn{6}{c}{SMD} & \multicolumn{6}{c}{MSL} & \multicolumn{6}{c}{SWaT}  & \multicolumn{6}{c}{PSM} \\
\cmidrule(lr){2-7}\cmidrule(lr){8-13}\cmidrule(lr){14-19} \cmidrule(lr){20-25}
Models & \rotatebox{90}{PATE} & \rotatebox{90}{PA-F1} & \rotatebox{90}{Standard-F1} & \rotatebox{90}{PA-AUC-ROC} & \rotatebox{90}{AUC-ROC} & \rotatebox{90}{VUS-ROC} &  \rotatebox{90}{PATE} & \rotatebox{90}{PA-F1} & \rotatebox{90}{Standard-F1} & \rotatebox{90}{PA-AUC-ROC} & \rotatebox{90}{AUC-ROC} & \rotatebox{90}{VUS-ROC} &  \rotatebox{90}{PATE} & \rotatebox{90}{PA-F1} & \rotatebox{90}{Standard-F1} & \rotatebox{90}{PA-AUC-ROC} & \rotatebox{90}{AUC-ROC} & \rotatebox{90}{VUS-ROC} &  \rotatebox{90}{PATE} & \rotatebox{90}{PA-F1} & \rotatebox{90}{Standard-F1} & \rotatebox{90}{PA-AUC-ROC} & \rotatebox{90}{AUC-ROC} & \rotatebox{90}{VUS-ROC}\\

\midrule
AnomalyTrans & $\textbf{0.06}$ & $0.91$ & $0.03$ & \textcolor{red}{$0.96$} & $0.49$ & $0.50$
             & $\textbf{0.13}$ & $0.94$ & $0.02$ & $0.97$ & $0.49$ & $0.52$
             & $\textbf{0.19}$ & $0.94$ & $0.02$ & $0.97$ & $0.53$ & $0.54$ 
             & $\textbf{0.33}$ & \textcolor{red}{$0.98$} & $0.02$ & \textcolor{red}{$0.99$} & $0.51$ & $0.52$ \\
             
DCDetector   & $\textbf{0.07}$ & $0.87$ & $0.01$ & $0.94$ & $0.50$ & $0.51$ 
             & $\textbf{0.14}$ & \textcolor{red}{$0.97$} & $0.02$ & \textcolor{red}{$0.98$} & $0.50$ & $0.58$
             & $\textbf{0.12}$ & \textcolor{red}{$0.96$} & $0.02$ & \textcolor{red}{$0.99$} & $0.49$ & $0.50$
             & $\textbf{0.32}$ & \textcolor{red}{$0.98$} & $0.02$ & \textcolor{red}{$0.99$} & $0.50$ & $0.52$\\        
             
USAD         & $\textbf{0.16}$ & \textcolor{red}{$0.94$} & $0.13$ & $0.91$ & $0.63$ & $0.72$
             & $\textbf{0.17}$ & $0.91$ & $0.06$ & $0.92$ & $0.53$ & $0.58$ 
             & \textcolor{darkergreen}{$\textbf{0.73}$} & $0.85$ & \textcolor{red}{$0.25$} & $0.83$ & \textcolor{red}{$0.82$} & \textcolor{red}{$0.61$}
             & $\textbf{0.45}$ & $0.89$ & $0.07$ & $0.91$ & $0.60$ & $0.61$\\        
             
LSTM         & $\textbf{0.25}$ & $0.80$ & \textcolor{red}{$0.14$} & $0.87$ & \textcolor{red}{$0.76$} & \textcolor{red}{$0.81$}
             & $\textbf{0.19}$ & $0.82$ & \textcolor{red}{$0.08$} & $0.87$ & $0.57$ & \textcolor{red}{$0.64$} 
             & $\textbf{0.71}$ & $0.82$ & $0.03$ & $0.85$ & \textcolor{red}{$0.82$} & $0.60$
             & $\textbf{0.55}$ & $0.93$ & \textcolor{red}{$0.15$} & $0.94$ & \textcolor{red}{$0.73$} & \textcolor{red}{$0.73$}\\        
             
Transformer  & \textcolor{darkergreen}{$\textbf{0.27}$} & $0.75$ & \textcolor{red}{$0.14$} & $0.84$ & $0.74$ & $0.80$
             & \textcolor{darkergreen}{$\textbf{0.20}$} & $0.40$ & $0.07$ & $0.63$ & \textcolor{red}{$0.60$} & $0.66$
             & $\textbf{0.72}$ & $0.82$ & $0.03$ & $0.85$ & \textcolor{red}{$0.82$} & $0.57$
             & \textcolor{darkergreen}{$\textbf{0.56}$} & $0.91$ & $0.14$ & $0.92$ & $0.72$ & $0.72$\\

\bottomrule
\end{tabular}
\vspace{1mm}

\centering
\begin{minipage}{0.47\linewidth}
    \includegraphics[width=\linewidth]{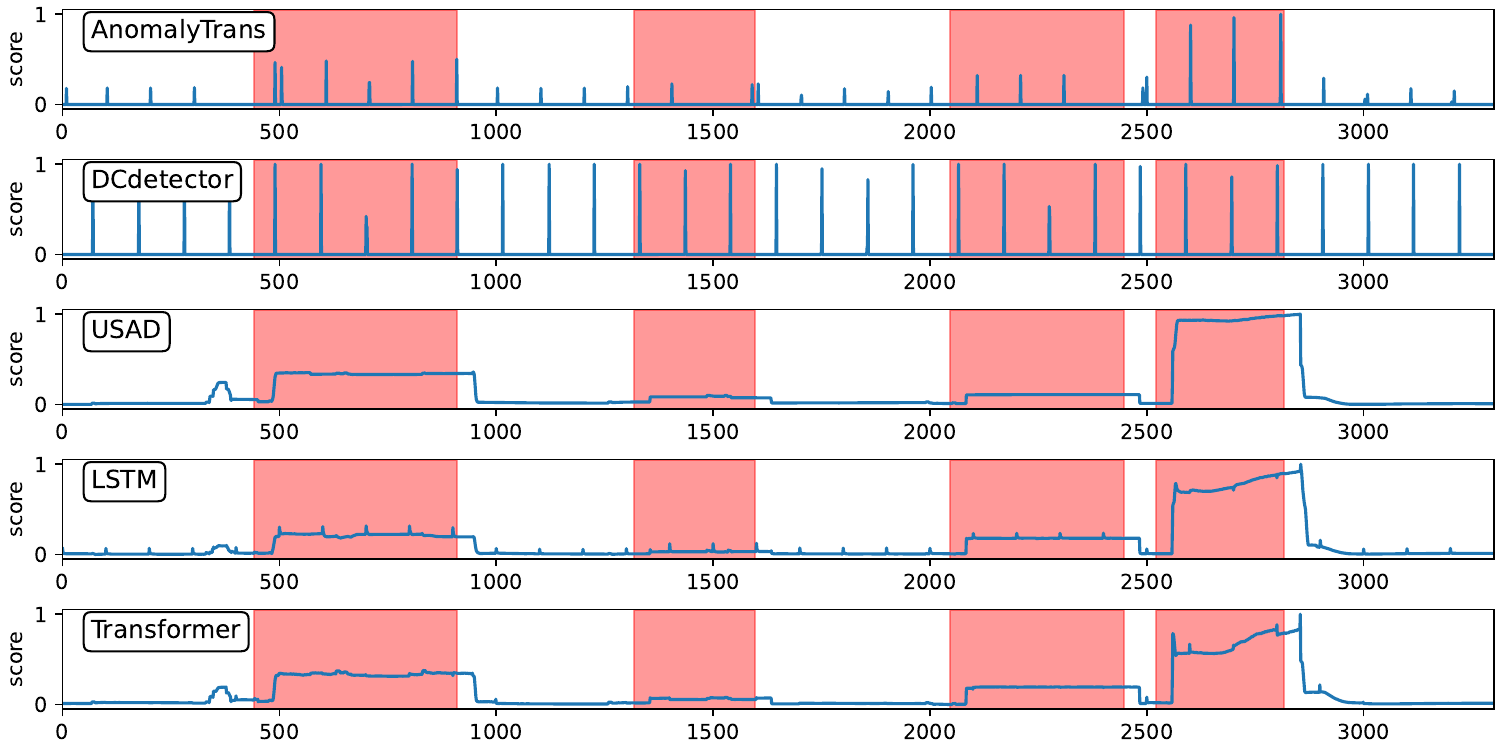}
    \centering
    \textit{(a) Anomaly Scores of SOTA models for SWaT dataset.}
    \label{fig:SWaT_SOTA_Example}
\end{minipage}\hfill % Add a small horizontal space
\begin{minipage}{0.47\linewidth}
    \includegraphics[width=\linewidth]{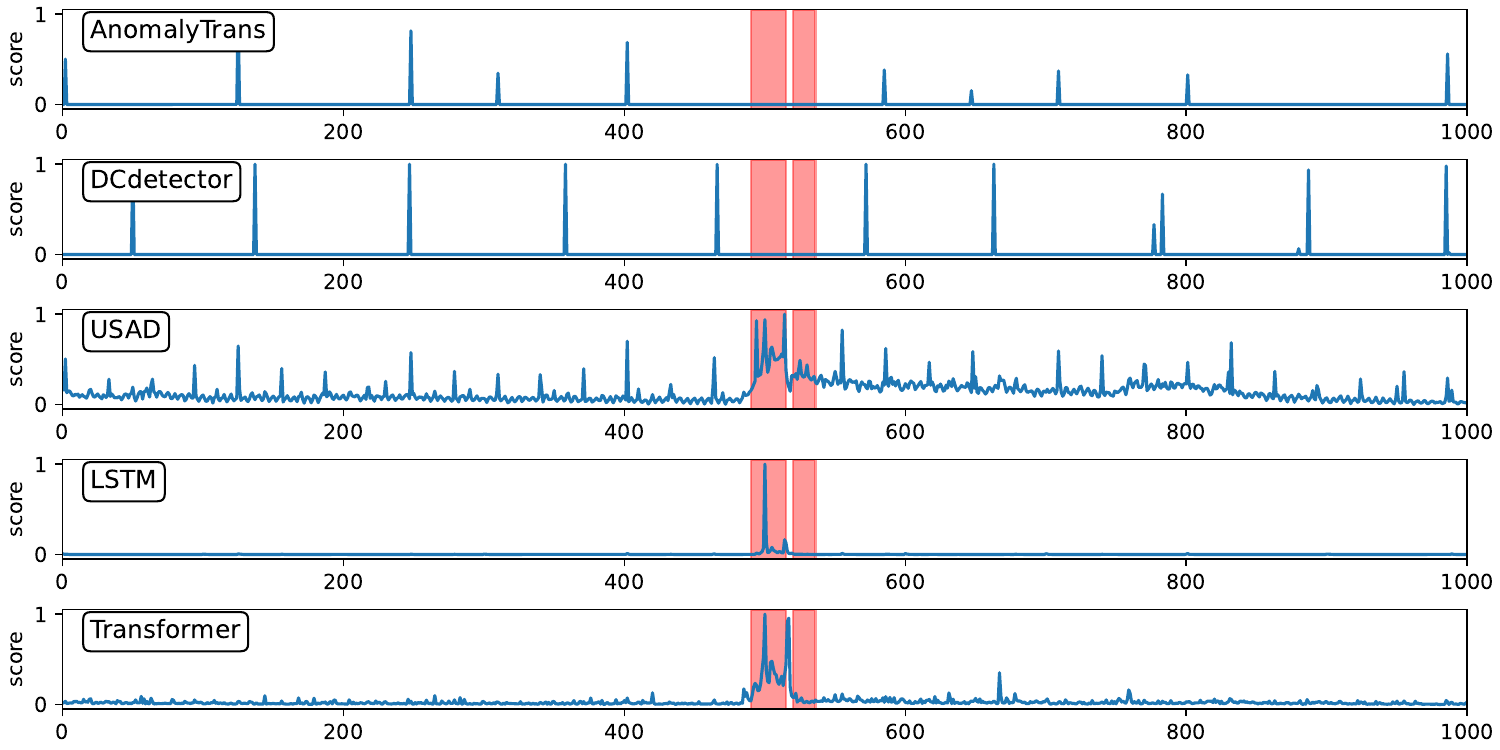}
    \centering
    \textit{(b) Anomaly Scores of SOTA models for SMD dataset.}
    \label{fig:SMD_SOTA_Example}
\end{minipage}
\vspace{2mm}
\captionof{figure}{\textit{Segments of anomaly scores of SOTA models for SWaT and SMD dataset. The highlighted regions in red indicate the true anomaly periods (labeled by an expert).}}
\label{fig:SOTA_Example}

\vspace{-5mm}
\end{figure*}

We analyzed the anomaly scores generated by 1) a Perfect Model, which serves as the benchmark by perfectly identifying anomalies; 2) established models like MultiVariate Normal distribution (MVN) \cite{MVN_Model}, Autoencoder (AE)\cite{AE_Model}, and Local Outlier Factor (LOF)\cite{LOF_Model}; 3) a baseline Random Score that assigns scores uniformly at random from a [0, 1] distribution. This selection covers a spectrum from theoretically ideal to practically random, offering a comprehensive view of the metrics' potential evaluation range. Detailed implementation of the models is available in our public code repository.

Figure \ref{fig:Real_Examples} showcases two real-world examples: (a) Weather Temperature data from UCR-KDD21 and (b) ECG data. The top row of each example shows the time series data with actual anomalies highlighted in red. The next rows illustrate the the output of the Perfect Model, and Models 1 and 2 (represented by MVN, LOF, or AE), demonstrating their respective detection scores. The final row displays a random score for baseline comparison. Table \ref{table:real-data-results} quantitatively compares various metrics. PATE consistently rates the Perfect Model highest and the Random Score lowest, showing its capability to recognize optimal detection and effectively penalize poor performance. In contrast, VUS-ROC/PR and AUC-ROC metrics seem less capable of such differentiation with the baselines. 

Moreover, PATE accurately takes into account the time series context and delayed detection effect, offering a more realistic and conservative assessment compared to VUS-ROC and AUC-ROC metrics, which appear to overestimate the performance of Models 1 and 2. This overestimation is evident in the Weather Temperature data, where Model 2 is inaccurately scored high by VUS-ROC and AUC-ROC despite its poor detection. Additionally, AUC-PR is also not sensitive in evaluation. For instance, in the Weather Temperature data, Model 1's delayed yet successful detection is incorrectly evaluated with a very low score, similar to the detection of Model 2. Similarly, in the ECG data, PATE's evaluation reflects the inconsistent anomaly detection pattern of Model 2 (AE) compared to Model 1 (MVN). However,  AUC-ROC/PR and VUS-ROC do not effectively consider this difference. Overall, PATE's assessments across both examples underscore its effectiveness in real-world applications.

\subsection{Impact Analysis: SOTA Models}

We re-evaluated several recent SOTA anomaly detection methods to not only assess their true performance but also to examine the stability of their ranking across various benchmark datasets when evaluated with different metrics, including PATE. Our comparative analysis includes models such as DCdetector~\cite{DCdetector}, AnomalyTrans~\cite{AnomalyTrans}, and USAD~\cite{USAD}, all of which have been recognized for their high performance in recent studies, alongside a Transformer and LSTM model, as simpler reconstruction-based anomaly detector baselines. These models are tested across the benchmark datasets of SMD~\cite{SMD}, MSL~\cite{MSL}, SWaT~\cite{SWaT}, and PSM~\cite{PSM}, used in previous works. Implementation details are available in our public code repository.

In the literature on SOTA models, the PA-F1 is the most frequently used and widely accepted metric. Additionally, in some cases, the standard F1 Score and Point-Adjusted variant of AUC-ROC (PA-AUC-ROC) are also employed. For a comprehensive comparison, we included these metrics in our comparative analysis. Results, shown in Table \ref{table:SOTA_Example}, highlight a significant discrepancy between PATE scores and those obtained from other metrics like PA-F1, Standard F1 Score, and PA-AUC-ROC. Notably, models that performed exceptionally well under PA-F1 and PA-AUC-ROC, such as AnomalyTrans and DCdetector, exhibit markedly lower scores when evaluated with PATE. For instance, for the SMD dataset, AnomalyTrans achieves a PA-F1 score of 0.91, showcasing high performance, yet its PATE score is only 0.06, indicating a substantial reduction in performance. To visually illustrate the differences in detection quality, Figure \ref{fig:SOTA_Example} shows a portion of the anomaly scores for the SWaT and SMD. The figures show that AnomalyTrans and DCdetector models struggle with consistent detection. In particular, for the SWaT, the peaky detections by these models hardly align with the expert-labeled anomaly intervals, and the high values reported for PA-F1 and PA-AUC-ROC do not reflect this detection pattern. This suggests that these metrics may overestimate model effectiveness. 

Next, Table \ref{table:SOTA_Example} shows that the Standard F1 Score, AUC-ROC, and VUS-ROC, do not exhibit such overestimations. However, they lack sensitivity to the finer aspects of detection as discussed in section \ref{se: Categories}. For instance, on the SWaT dataset, the Standard F1 Score is not able to distinguish between the good performing LSTM and Transformer and the poorly performing AnomalyTrans and DCdetector, see also Figure \ref{fig:SOTA_Example}~(a). Furthermore, AUC-ROC does not reflect the small differences between USAD, LSTM, or Transformer. The scores of this metric suggest that all models have an identical performance, that does not match the reality of their output. Moreover, while VUS-ROC offers a slightly better distinction among models than AUC-ROC, its limited scoring range (e.g., 0.54 for AnomalyTrans and 0.57 for Transformer) makes it challenging to clearly identify models that perform exceptionally well from those that do not. Meanwhile, PATE offers a more consistent and transparent assessment. It can be seen that PATE gives a relatively higher score to USAD (0.73), Transformer (0.72), and LSTM (0.71) according to their better detection pattern. PATE even slightly prefers USAD over LSTM, although the difference is small.

\begin{figure}[H]
\centering
\includegraphics[width=\columnwidth]{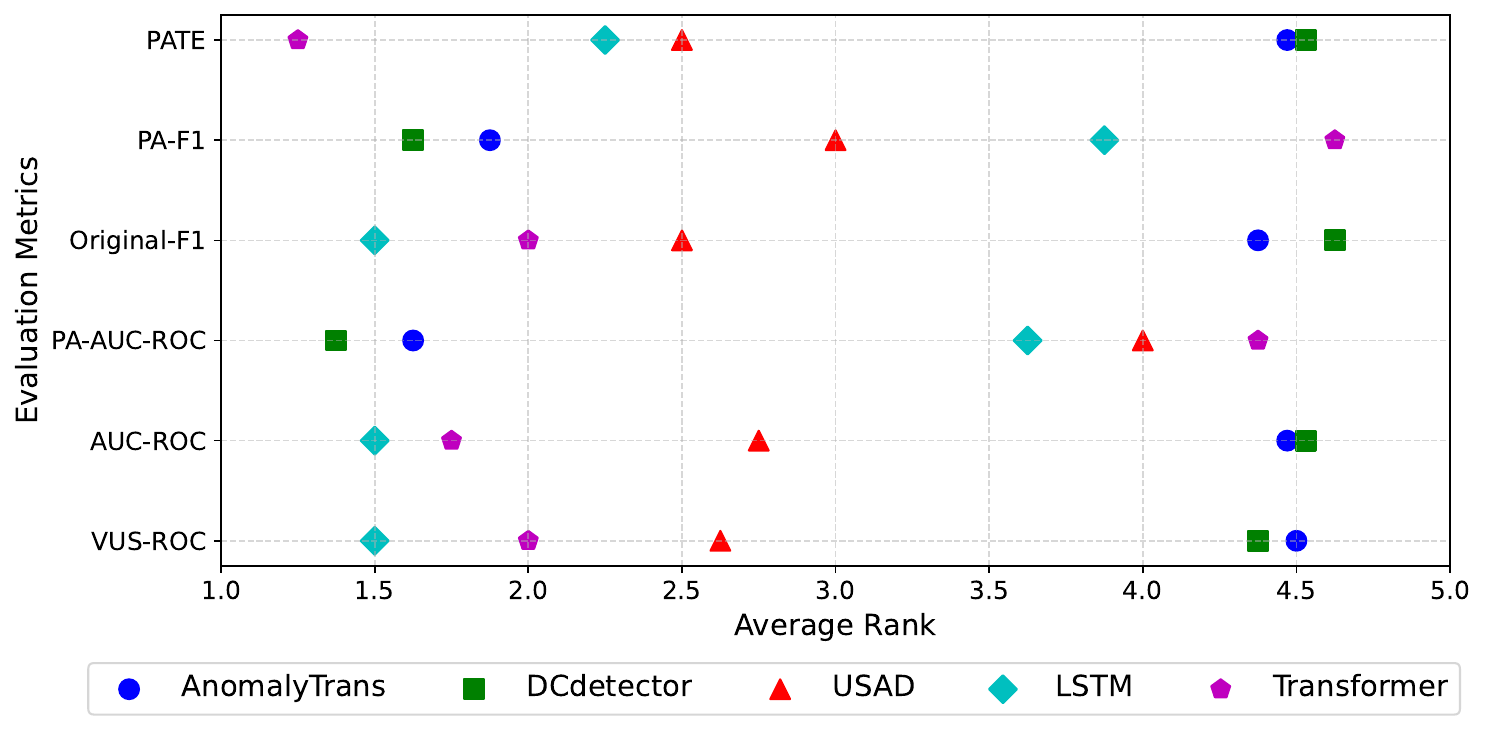}
\vspace{-5mm}
\caption{\textit{Average rankings of different models for various evaluation metrics across all benchmark datasets.}}
\label{fig:Rankings_analysis}
\vspace{-2mm}
\end{figure}

We also explored the average rankings of the models for all metrics across all four benchmark datasets. Figure \ref{fig:Rankings_analysis} presents these rankings, highlighting noticeable differences in the standings of the models when using different metrics. The average rankings based on the PA-F1 metric place DCdetector at the forefront with an average rank of $1.62$, followed by AnomalyTrans ($1.88$), USAD ($3.00$), LSTM ($3.88$), and Transformer ($4.62$). However, when evaluated with PATE, a significant shift occurs: Transformer and LSTM emerge as the top-performing models with ranks of $1.38$ and $2.12$, respectively, while AnomalyTrans and DCdetector drop to the bottom ranks of $4.50$ each. This variance underscores the critical impact of the chosen evaluation metric and the importance of selecting a proper metric such as PATE.

\vspace{-2.5mm}

\section{Ablation Analysis: Buffer Sizes}

\begin{figure}[!h]
\vspace{-3.7mm}
\centering
\includegraphics[width=\columnwidth, keepaspectratio]{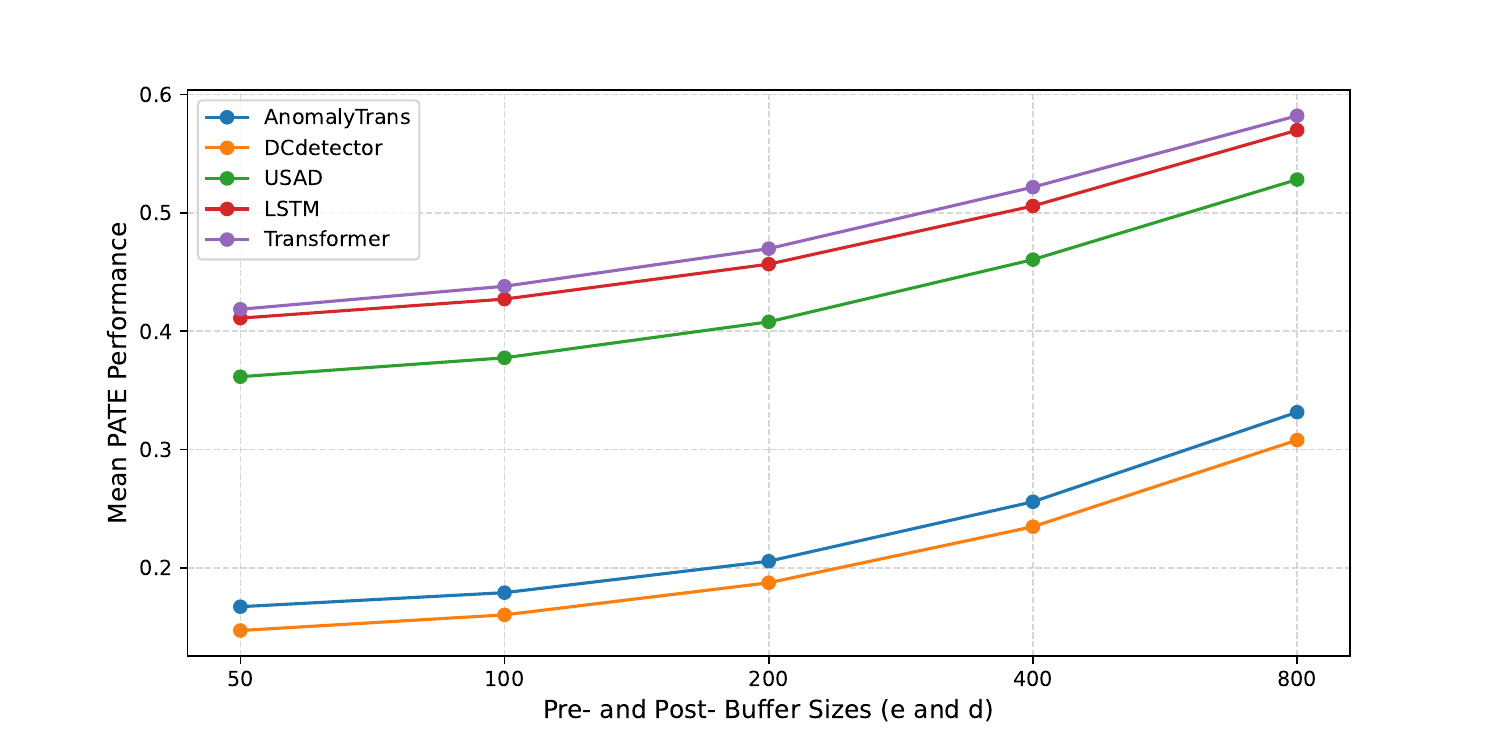}
\vspace{-8mm}
\caption{\textit{Mean PATE performance of all models across all datasets for different Pre and Post-Buffer sizes (e = d).}}
\label{fig:PATE_Ablation_Analysis}
\vspace{-4mm}
\end{figure}

The adaptability of PATE to accommodate different buffer sizes is one of its key strengths. This flexibility allows for an expert-driven and context-specific approach to model evaluation, ensuring that the unique characteristics of each dataset are appropriately considered. Figure \ref{fig:PATE_Ablation_Analysis} illustrates the mean performance of DCdetector, AnomalyTrans, USAD, LSTM, and Transformer across all four benchmark datasets using PATE. Results show that PATE consistently ranks models such as Transformer and LSTM the highest across different buffer sizes. This consistency in model rankings, irrespective of buffer size, highlights PATE's robustness as an evaluation metric, and showcases PATE's reliability for diverse applications, ensuring a consistent and dependable assessment for anomaly detection models.

\section{Discussion and Conclusion}

We proposed PATE, a novel approach to evaluate anomaly detection models in time series data. PATE addresses the limitations of existing evaluation metrics by categorizing the anomaly and prediction events and assigning proximity-based weighting, considering different buffer zones around the anomaly event. PATE computes the area under the Precision-Recall curve, where the Precision and Recall are computed from weighted versions of True Positive, False Positive, and False Negative performances. 

Our experiments with both synthetic and real-world data demonstrate that PATE effectively differentiates between models based on their actual performance, considering early and delayed detection, onset response time, coverage level of the anomaly event, and consistency in detection. The re-evaluation of SOTA anomaly detection methods using PATE reveals notable differences in performance assessments compared to other metrics. For instance, point-adjusted metrics often overestimate the performance of models. However, in practice, metrics such as ROC-AUC and VUS-ROC offer more reasonable estimates for SOTA models, though they might overlook subtle detection errors and sometimes lack discriminability between models. This analysis not only questions the true performance of current SOTA models but also indicates a shift in their rankings, challenging the prevailing understanding of the superiority of these models. PATE's ability to provide a more matching, context-sensitive, and transparent assessment highlights its potential as a more appropriate metric that can set a new standard for evaluating advancements in anomaly detection. Additionally, PATE's adaptability to various buffer sizes without compromising consistency and fairness in model evaluation further highlights its robustness and applicability across diverse applications. 

To address the specific scenarios where either an expert has predetermined the threshold or models inherently output binary labels, we have developed \textit{PATE-F1} as an essential extension of the original PATE framework. The methodology and experimental insights on \textit{PATE-F1} are detailed in Appendix~\ref{PATE_F1}. \textit{PATE-F1} effectively distinguishes between different scenarios based on temporal proximity, duration, coverage level, and response timing, setting it apart from other metrics that face limitations in capturing these aspects in evaluation. Additionally, our findings indicate that the original PATE framework, through strategic threshold application, naturally extends to effectively evaluate binary outputs. However, employing \textit{PATE-F1} in such scenarios offers a more direct and simplified approach. This adaptation ensures PATE's methodology remains a versatile and applicable measure across a broader spectrum of anomaly detection approaches and contexts. 

In conclusion, PATE represents a significant advancement in the evaluation of time series anomaly detection methods which has the potential to guide future research, influence industry adoption, and enhance the development of practical applications in critical domains such as healthcare and finance.

\begin{acks}
Funding: This work was supported by the Dutch Research Council (NWO) [grant numbers 628.011.214].
\end{acks}

\bibliographystyle{ACM-Reference-Format}
\bibliography{sample-base}

%%%%%%%%%%%%%%%%%%%%%%%%%%%%%%%%%%%%%%%%%%%%%%%%%%%%%%%%%%%%%%%%%%%%%%%%%%%%%%%
%%%%%%%%%%%%%%%%%%%%%%%%%%%%%%%%%%%%%%%%%%%%%%%%%%%%%%%%%%%%%%%%%%%%%%%%%%%%%%%
% APPENDIX
%%%%%%%%%%%%%%%%%%%%%%%%%%%%%%%%%%%%%%%%%%%%%%%%%%%%%%%%%%%%%%%%%%%%%%%%%%%%%%%
%%%%%%%%%%%%%%%%%%%%%%%%%%%%%%%%%%%%%%%%%%%%%%%%%%%%%%%%%%%%%%%%%%%%%%%%%%%%%%%
%\newpage
\appendix

\newpage

\section{Reproducibility Statement}

To ensure the reproducibility of our work, the source code, along with comprehensive documentation, is publicly available at:

\noindent\url{https://github.com/Raminghorbanii/PATE}.

This repository includes detailed instructions for using PATE, including how to set the buffer size, and complete descriptions of all models implemented for our experiments, covering configuration settings, training procedures, and experimental details to ensure accurate replication. Researchers seeking additional information are encouraged to contact the corresponding author.

%%%%%%%%%%%%%%%%%%%%%%%%%%%%%%%%%%%%%%%%
%%%%%%%%%%%%%%%%%%%%%%%%%%%%%%%%%%%%%%%%
%%%%%%%%%%%%%%%%%%%%%%%%%%%%%%%%%%%%%%%%

\section{Effect of Anomaly Length on Buffer Weights}
\label{Effect_lenght_Ws}

\begin{figure}[!h]
\centering
\includegraphics[width=\columnwidth]{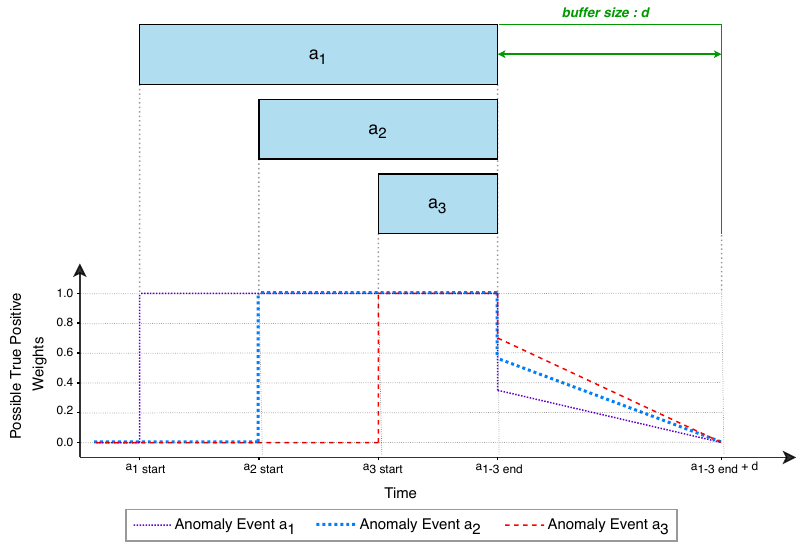} \
\vspace{-4.5mm}

\caption{\textit{Potential True Positive (TP) weights relative to the anomaly events with varying lengths}. The graph illustrates the higher TP weights for detections following the shortest anomaly event \(a_3\), and the progressively lower weights for the longer events \(a_1\) and \(a_2\).}

\label{fig:Ablation on Ws of Post-Buffer}
\end{figure}

To explore the effect of anomaly length on the assignment of weights within the PATE framework, we consider three distinct anomaly events with varying durations: \(a_1\), \(a_2\), and \(a_3\), with \(a_1\) being the longest and \(a_3\) the shortest. Each was followed by a post-buffer zone of fixed size \(d\). Figure~\ref{fig:Ablation on Ws of Post-Buffer} depicts the potential True Positive (TP) weights along the timeline, capturing the period before the anomaly, within its range, and throughout the post-buffer zone. The analysis of this figure indicates that TP weights for detections in the post-buffer zone are higher for \(a_3\), the shortest anomaly, and progressively lower for \(a_1\) and \(a_2\), the longer anomalies. This observation underscores the direct correlation between the duration of an anomaly and the corresponding TP weights assigned to post-buffer detections. Higher TP weights for detections following shorter anomalies signify the critical nature of these detections, as they are in closer proximity to the anomaly onset. The PATE weighting mechanism accommodates this by adjusting the weights based on the distance from detections to the entire anomaly. This phenomenon also extends to the pre-buffer zone, where early detections are similarly influenced by the length of the forthcoming anomaly. 

\vspace{-4mm}

%%%%%%%%%%%%%%%%%%%%%%%%%%%%%%%%%%%%%%%%
%%%%%%%%%%%%%%%%%%%%%%%%%%%%%%%%%%%%%%%%
%%%%%%%%%%%%%%%%%%%%%%%%%%%%%%%%%%%%%%%%

\section{Clarification on early and delayed detections }
\label{Clarification_DD_ED}
To understand the distinct approaches PATE takes toward Early Detection (in the pre-buffer zone) and Delayed Detection (in the post-buffer zone), it is essential to consider the foundational goal of this evaluation metric.

For an anomaly detector, the ability to learn from past data and accurately predict future anomalies is essential. An early prediction that fails to correspond with an actual, subsequent anomaly suggests a fundamental modeling failure of the data's underlying structure—like sounding an alarm for an event that never happens. Ideally, if a model detects early signs of an impending anomaly, it should also identify the anomaly when it occurs. The early signs—small changes or patterns of deterioration—lead to a larger and more evident departure from the norm. If the model has correctly identified these early signs, it should also recognize the anomaly itself, given the now more noticeable deviation. When the early detection is successfully followed by a true detection of the anomaly, the early detection is not considered just a lucky guess. It supports the model's predictive power and consistency.

In contrast, the context for delayed detection significantly differs as it showcases the capability of the model to identify anomalies post hoc. The model is apparently able to detect some deviation in the input, albeit a bit late. Such late detections still allow for the identification of the anomaly. Failing to have True Positive detections in the anomaly event is therefore not considered fatal for the Delayed Detection.

\begin{figure}[H]
\centering
\vspace{-2mm}
\includegraphics[width=0.95\columnwidth]{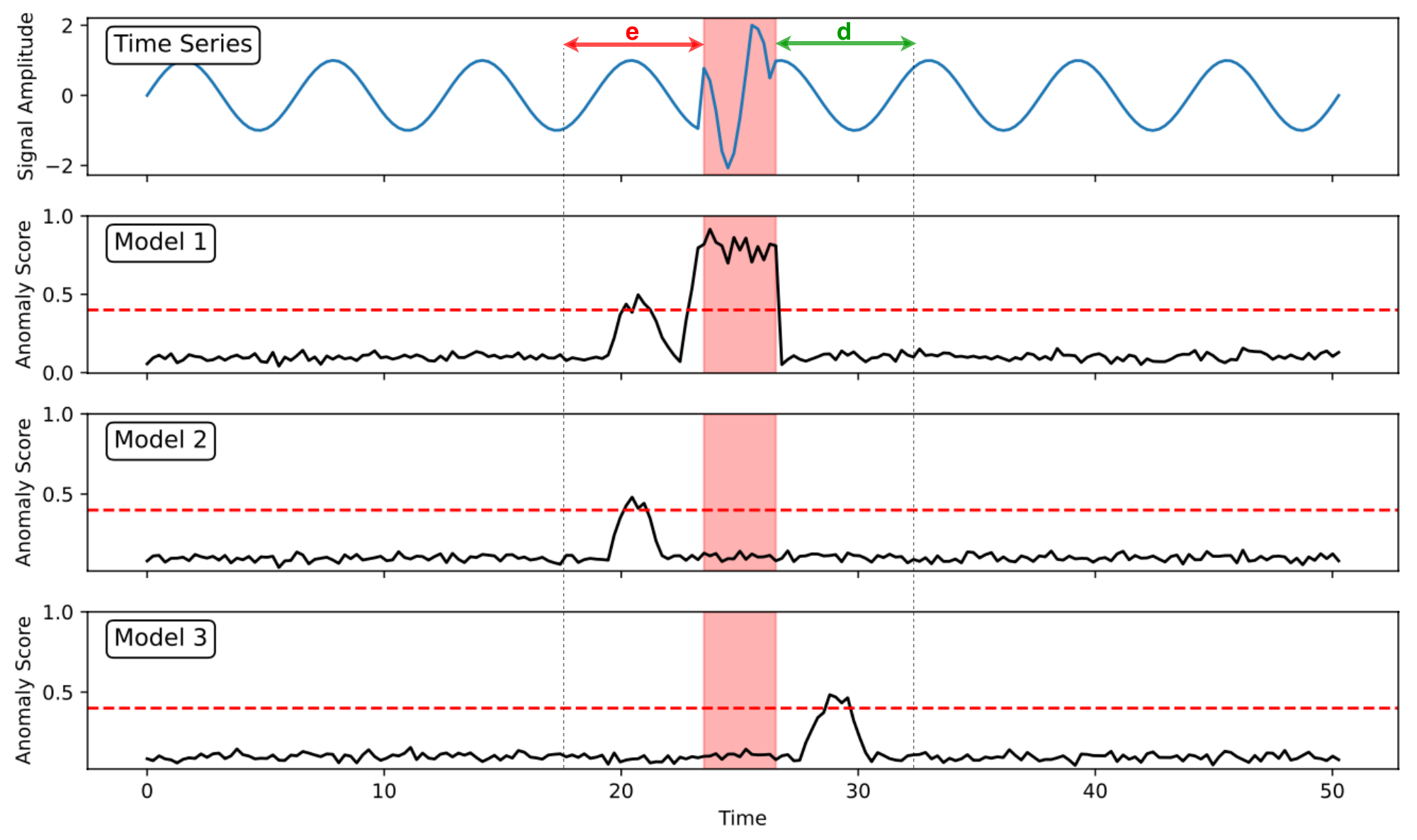} \
\vspace{-3mm}
\caption{\textit{Comparative evaluation of model responses to an anomalous event in time series data.}}

\label{fig:Clarification_EE_DD}
\vspace{-2mm}

\end{figure}

Figure~\ref{fig:Clarification_EE_DD} shows the detection responses by three different models to an anomalous event, shown by the shaded area in red. Model 1 (top panel) reveals an early detection followed by True Positive detections, indicated by peaks aligning with the anomaly window. This pattern exemplifies an acceptable detection where the model preemptively and accurately identifies an anomaly. Model 2 (middle panel), however, demonstrates early detection without subsequent TPs during the actual anomaly, missing the critical deviation. This outcome might suggest a misinterpretation of the anomaly pattern by Model 2, potentially leading to a false alarm scenario. Conversely, Model 3 (bottom panel) shows a peak that arises post the onset of the anomaly, exemplifying a delayed detection. This detection is valued as it demonstrates the capacity of the model for retrospective analysis, acknowledging and learning from the anomaly event after its occurrence. 

%%%%%%%%%%%%%%%%%%%%%%%%%%%%%%%%%%%%%%%%
%%%%%%%%%%%%%%%%%%%%%%%%%%%%%%%%%%%%%%%%
%%%%%%%%%%%%%%%%%%%%%%%%%%%%%%%%%%%%%%%%

\section{PATE-F1 - Adjusted for binary scores}
\label{PATE_F1}
\vspace{2mm}

\textbullet\, \textbf{\textit{Methodology:}}
To enhance the applicability of PATE in scenarios where models use predetermined thresholds or where expert knowledge informs threshold determination, we propose an adapted version, \textit{PATE-F1}. This adaptation leverages the core principles of PATE by assigning proximity-specific weights to categorized points and calculating weighted Precision and Recall. Unlike the original PATE, which evaluates a range of thresholds (\(\theta\)), PATE-F1 is tailored for binary scenarios, without the variation of thresholds but rather different combinations of buffer zones (\(e\) and \(d\)). For each combination, weighted Precision and Recall are calculated using equations \ref{eq:precision} and \ref{eq:recall} as detailed in Section \ref{se:PATE}. Subsequently, the F1 score for each combination is determined as follows:

\begin{equation}
\text{F1-Score}_{e,d} = 2 \times \frac{\text{Precision}_{e,d} \times \text{Recall}_{e,d}}{\text{Precision}_{e,d} + \text{Recall}_{e,d}}
\end{equation}

\

The overall PATE-F1 score is then computed as the average of these F1 scores across all buffer zone combinations:

\begin{equation}
\text{PATE-F1} = \frac{1}{|E| \times |D|} \sum_{e \in E, d \in D} \text{F1-Score}_{e,d}
\end{equation}

\

Here, \(|E|\) and \(|D|\) represent the number of distinct pre-buffer (\(e\)) and post-buffer (\(d\)) sizes, respectively. 

\

\begin{figure}[!ht]
\centering
\includegraphics[width=\columnwidth]{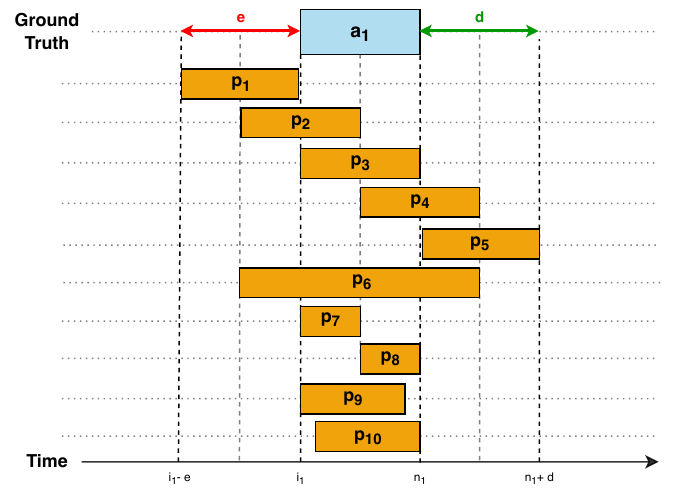}
\vspace{-7mm}
\caption{\textit{Examples with synthetic prediction events (binary scores)}. The figure shows the placement of different prediction events $p_l(\theta)$ from a binary anomaly detector.}
\label{fig:scenarios_PATEF1}
\end{figure}

\noindent\textbullet\, \textbf{\textit{Experimental Results:}}
We extend our analysis to \textit{PATE-F1} by comparing the evaluations against threshold-dependent metrics, tailored for binary score predictions. Figure~\ref{fig:scenarios_PATEF1} shows 10 different detection scenarios shown by prediction events $p_{1}, \dots, p_{10}$. Table~\ref{table:Results-F1PATE} shows that similar to the original PATE, \textit{PATE-F1} effectively differentiates between scenarios based on temporal proximity, duration, coverage level, and response timing. This alignment with PATE's evaluation logic underlines the adaptability of our methodology to binary score scenarios without compromising the depth of analysis provided by the range of thresholds in the original framework.

\begin{table}
\captionof{table}{\textit{Comparison of evaluation metrics for synthetic prediction event examples depicted in Figure~\ref{fig:scenarios_PATEF1}}. 'F1' refers to the F1 Score.}
\vspace{-3mm}
\label{table:Results-F1PATE}
\fontsize{6pt}{7pt}\selectfont % Custom font size
%\setlength{\tabcolsep}{4pt} % Adjust the space between columns
%\small % or \footnotesize or \scriptsize
\resizebox{\columnwidth}{!}{
\begin{tabular}{c|cc|ccccc}
\toprule

&\multicolumn{7}{c}{\begin{tabular}{@{}c@{}}  Metrics \end{tabular}} \\
\cmidrule(lr){2-8} 
\rotatebox{90}{Scenarios} & \rotatebox{90}{PATE} & \rotatebox{90}{PATE-F1} & \rotatebox{90}{Standard-F1} & \rotatebox{90}{PA-F1} & \rotatebox{90}{R-based-F1} & \rotatebox{90}{ETS-Aware-F1} & \rotatebox{90}{Affiliation-F1} \\
\midrule

$p_{1}$ & $\textbf{0.03}$ & $\textbf{0.00}$ & $0.00$ & $0.00$ & $0.00$ & $0.00$ & $0.94$ \\
$p_{2}$ & $\textbf{0.76}$ & $\textbf{0.75}$ & $0.50$ & $0.80$ & $0.60$ & $0.75$ & $0.98$ \\
$p_{3}$ & $\textbf{1.00}$ & $\textbf{1.00}$ & $1.00$ & $1.00$ & $1.00$ & $1.00$ & $1.00$ \\
$p_{4}$ & $\textbf{0.69}$ & $\textbf{0.66}$ & $0.50$ & $0.80$ & $0.60$ & $0.75$ & $0.98$\\
$p_{5}$ & $\textbf{0.31}$ & $\textbf{0.28}$ & $0.00$ & $0.00$ & $0.00$ & $0.00$ & $0.94$ \\
$p_{6}$ & $\textbf{0.87}$  & $\textbf{0.85}$ & $0.67$ & $0.67$ & $0.75$ & $0.86$ & $0.98$\\
$p_{7}$ & $\textbf{0.85}$ & $\textbf{0.81}$ & $0.67$ & $1.00$ & $0.75$ & $0.86$ & $0.99$\\
$p_{8}$ & $\textbf{0.77}$  & $\textbf{0.67}$ & $0.67$ & $1.00$ & $0.75$ & $0.86$ & $0.99$ \\
$p_{9}$ & $\textbf{0.95}$  & $\textbf{0.95}$ & $0.86$ & $1.00$ & $0.89$ & $0.93$ & $1.00$ \\
$p_{10}$ & $\textbf{0.88}$  & $\textbf{0.86}$ & $0.86$ & $1.00$ & $0.89$ & $0.93$ & $1.00$ \\

\bottomrule
\end{tabular}
}

\vspace{-2mm}
\end{table}

%%%%%%%%%%%%%%%%%%%%%%%%%%%%%%%%%%%%%%%%
%%%%%%%%%%%%%%%%%%%%%%%%%%%%%%%%%%%%%%%%
%%%%%%%%%%%%%%%%%%%%%%%%%%%%%%%%%%%%%%%%

\section{Complexity Time Analysis}

We evaluated the computational efficiency of the PATE algorithm against established metrics like AUC-PR and VUS-PR through experiments on synthetic and real benchmark datasets. These experiments were conducted on a standard MacBook with a 2 GHz Quad-Core Intel Core i5 processor, Intel Iris Plus Graphics 1536 MB, and 16 GB RAM, reflecting the performance on commonly available hardware. Although PATE supports parallel execution to potentially decrease computation time, especially on High-Performance Computing (HPC) systems, we used a serial computation approach for consistent comparisons with other metrics.

\

\begin{figure}[H]
\centering
\includegraphics[width=0.95\columnwidth]{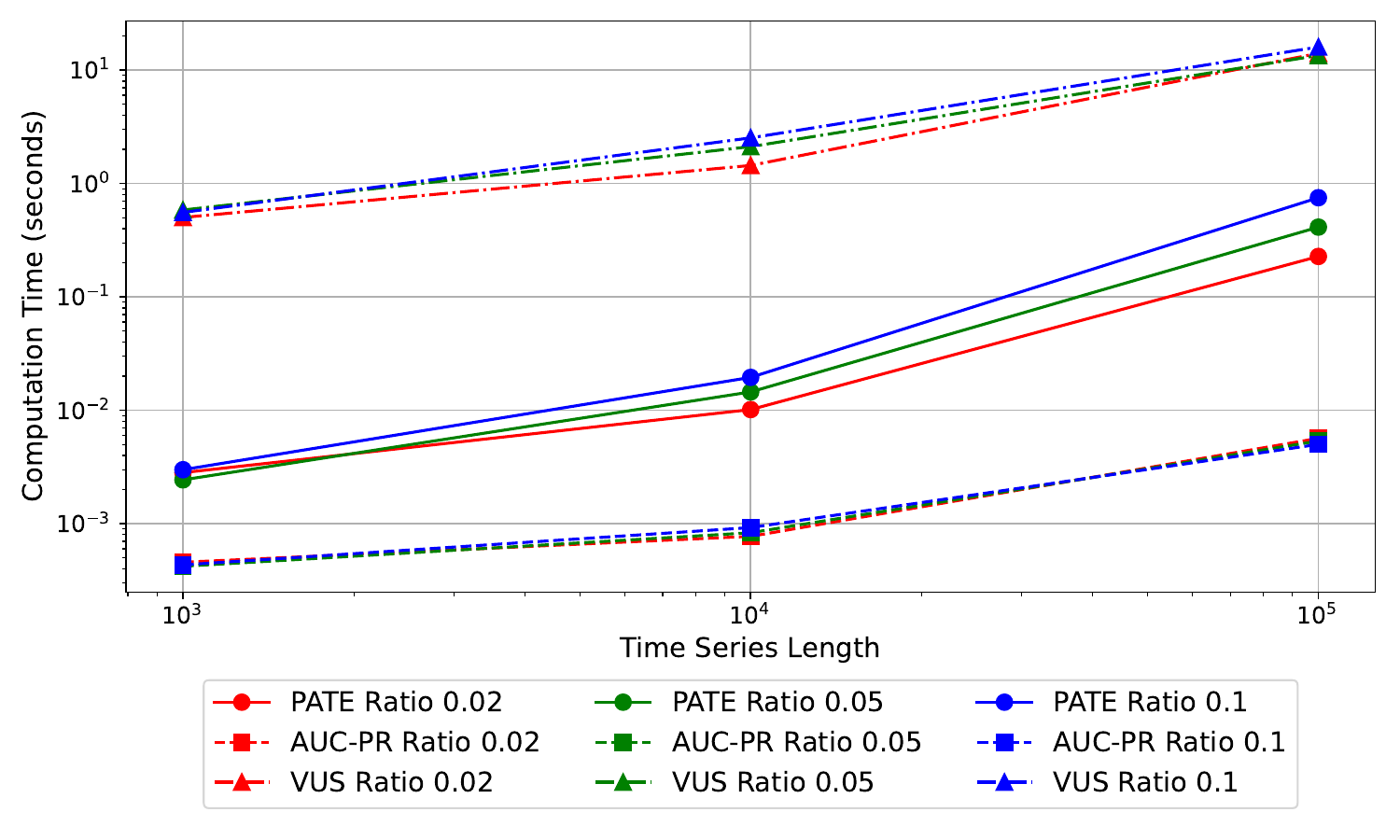}
\vspace{-4mm}
\caption{\textit{Computation time of PATE on synthetic data with varying lengths and anomaly ratios.}}
\label{fig:time_complexity}
\end{figure}

\textbullet\, \textbf{\textit{Synthetic Data Experiments:}}
We generated synthetic time series data ranging from 1,000 to 100,000 points with anomaly ratios of 2\%, 5\%, and 10\% to reflect various common scenarios. As shown in Figure~\ref{fig:time_complexity}, PATE’s computation time increases linearly with data length and varies slightly with different anomaly ratios. Despite this, computation times remained under one second across all conditions, highlighting PATE’s efficiency without parallel processing.

\vspace{-4mm}

\textbullet\, \textbf{\textit{Benchmark Dataset Experiments:}}
We validated PATE on all standard benchmark datasets used in this study. As shown in Table~\ref{table:metric-comparison_TimeComplexity}, PATE’s computation times are comparable to those of the AUC-PR metric and significantly faster than the VUS metric, remaining under one second for smaller datasets and under two seconds for larger ones. Note that further speed enhancements could be achieved on HPC systems or with parallel processing.

\begin{table}[H]
\caption{\textit{Computation times (in seconds) for evaluation metrics across benchmark datasets.}}

\vspace{-2.5mm}
\label{table:metric-comparison_TimeComplexity}
\fontsize{5.5pt}{6pt}\selectfont % Custom font size
%\setlength{\tabcolsep}{4pt} % Adjust the space between columns
%\small % or \footnotesize or \scriptsize
\resizebox{\columnwidth}{!}{
\begin{tabular}{c|cc|ccc}
\toprule

& & &\multicolumn{3}{c}{Evaluation Metrics} \\
\cmidrule(lr){4-6} 
\rotatebox{90}{Datasets} & \rotatebox{90}{\begin{tabular}{@{}c@{}} Time series \\ Size \end{tabular}} & \rotatebox{90}{\begin{tabular}{@{}c@{}} Anomaly \\ Ratio \end{tabular}} & \rotatebox{90}{AUC-PR} & \rotatebox{90}{VUS-PR} & \rotatebox{90}{PATE} \\
\midrule

MSL & $73700$ & $10 \%$ & $0.007$ & $42.315$ & $0.278$   \\
PSM & $87800$ & $4 \%$ & $0.013$ & $51.683$ & $0.634$   \\
SWaT & $449900$ & $12 \%$ & $0.267$ & $249.573$ & $1.895$   \\
SMD & $708400$ & $4 \%$ & $0.064$ & $462.252$ & $1.796$   \\

\bottomrule
\end{tabular}
}
\end{table}

%%%%%%%%%%%%%%%%%%%%%%%%%%%%%%%%%%%%%%%%
%%%%%%%%%%%%%%%%%%%%%%%%%%%%%%%%%%%%%%%%
%%%%%%%%%%%%%%%%%%%%%%%%%%%%%%%%%%%%%%%%

%%%%%%%%%%%%%%%%%%%%%%%%%%%%%%%%%%%%%%%%%%%%%%%%%%%%%%%%%%%%%%%%%%%%%%%%%%%%%%%
%%%%%%%%%%%%%%%%%%%%%%%%%%%%%%%%%%%%%%%%%%%%%%%%%%%%%%%%%%%%%%%%%%%%%%%%%%%%%%%

\end{document}